\documentclass[9.5pt,journal,compsoc]{IEEEtran}

\usepackage[nocompress]{cite}
\usepackage{multicol}
\usepackage{multirow}
\usepackage{makecell}
\usepackage{booktabs}
\usepackage{amsmath}
\usepackage{amssymb}
\usepackage{amsfonts}
\usepackage{enumitem}
\usepackage{bm}
\usepackage{graphicx}
\usepackage{subfigure}
\usepackage{soul}
\usepackage{tikz}
\usepackage{tikz-3dplot}
\usepackage{tabularx}
\usepackage{ragged2e}
\usepackage{xr}
\usepackage{xr-hyper}
\usepackage{totcount}

\regtotcounter{figure}
\regtotcounter{table}
\usetikzlibrary{calc}
\usetikzlibrary{shapes.geometric}
\usetikzlibrary{shapes.arrows}
\usetikzlibrary{positioning}
\usetikzlibrary{shapes.multipart}
\usetikzlibrary{decorations.pathreplacing} %

\def\eg{\emph{e.g.}}
\def\ie{\emph{i.e.}}
\def\lseg{\ddot{l}}
\def\line{{\bm l}}
\makeatletter
\def\@IEEEsectpunct{\ \,}
\def\paragraph{\@startsection{paragraph}{4}{\z@}{0.5ex plus 1.5ex minus 0.5ex}%
	{0ex}{\normalfont\normalsize\sffamily\bfseries}}
\makeatother
\makeatletter
\def\@IEEEsectpunct{\ \,}
\def\paragrapha{\@startsection{paragraph}{4}{\z@}{1.5ex plus 1.5ex minus 0.5ex}%
	{0ex}{\normalfont\normalsize\sffamily\bfseries}}
\makeatother

\usepackage{url}

\begin{document}

\title{Holistically-Attracted Wireframe Parsing:\\ From Supervised to Self-Supervised Learning}

\author{Nan Xue,
        Tianfu Wu,
        Song Bai,
        Fu-Dong Wang,
        Gui-Song Xia,
        Liangpei Zhang,
        Philip H.S. Torr
\IEEEcompsocitemizethanks{
\IEEEcompsocthanksitem N. Xue is with Ant Group.
\IEEEcompsocthanksitem T. Wu is with the Department of ECE, North Carolina State University.
\IEEEcompsocthanksitem S. Bai is with ByteDance AI Lab.
\IEEEcompsocthanksitem F.-D. Wang is with Ant Group.
\IEEEcompsocthanksitem G.-S. Xia and L. Zhang are with Wuhan University.
\IEEEcompsocthanksitem P. Torr is with the University of Oxford.\\
}
}
\markboth{IEEE Trans. on Pattern Analysis and Machine Intelligence}{Manuscript}

\IEEEtitleabstractindextext{%
\justifying
\begin{abstract}
This article presents Holistically-Attracted Wireframe Parsing (HAWP), a method for geometric analysis of 2D images containing wireframes formed by line segments and junctions. HAWP utilizes a parsimonious Holistic Attraction (HAT) field representation that encodes line segments using a closed-form 4D geometric vector field. The proposed HAWP consists of three sequential components empowered by end-to-end and HAT-driven designs: (1) generating a dense set of line segments from HAT fields and endpoint proposals from heatmaps, (2) binding the dense line segments to sparse endpoint proposals to produce initial wireframes, and (3) filtering false positive proposals through a novel endpoint-decoupled line-of-interest aligning (EPD LOIAlign) module that captures the co-occurrence between endpoint proposals and HAT fields for better verification. Thanks to our novel designs, HAWPv2 shows strong performance in fully supervised learning, while HAWPv3 excels in self-supervised learning, achieving superior repeatability scores and efficient training (24 GPU hours on a single GPU). Furthermore, HAWPv3 exhibits a promising potential for wireframe parsing in out-of-distribution images without providing ground truth labels of wireframes.
\end{abstract}

}

\maketitle

\IEEEdisplaynontitleabstractindextext

\IEEEpeerreviewmaketitle

\IEEEraisesectionheading{\section{Introduction}\label{sec:introduction}}

\IEEEPARstart{D}{epicting} image contents with geometric entities/patterns such as salient points, line segments, and planes/surfaces has been shown as an effective encoding scheme of visual information evolved in primate visual systems, which in turn has long motivated the computer vision community to make tremendous efforts on computing the primal sketch~\cite{MarrH80} of natural images consisting of different forms including, but not limited to, the blobs~\cite{SIFT,MSER,MikolajczykS04}, corners/junctions~\cite{HarrisS88,MikolajczykS04,XueXBZS18,XiaDG14}, edges~\cite{bsds,HED-ijcv,CATS}, and line segments~\cite{Ballard81,VonGioi2010} since the 1960s. 
Modeling and computing the primal sketches have remained a long-standing problem, and it plays important roles in many downstream tasks including 3D reconstruction~\cite{0009LYZ21,HoW3D,PlaneTR,Gomez-OjedaMZSJ19,LiMZF0S020} and scene parsing~\cite{wu2010learning,DuanL15}, as well as high-level visual recognition tasks~\cite{LazarowXT22,XiaHXLZ19,abs-2208-00609}.

\begin{figure}
\vspace{-4em}
\centering
\subfigure[HAWPv1 on the Wireframe Dataset (FSL Model)\label{fig:teaser-a}]{
\includegraphics[width=0.28\linewidth]{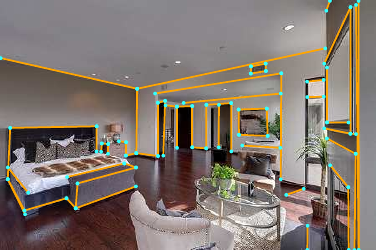}
\includegraphics[width=0.28\linewidth]{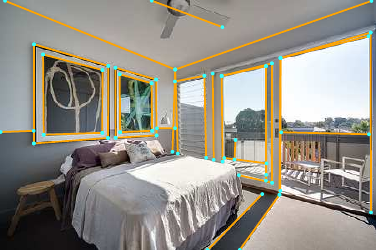}
\includegraphics[width=0.28\linewidth]{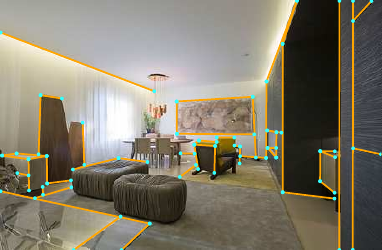}
}\vspace{-3mm}

\subfigure[HAWPv2 on the Wireframe Dataset (FSL Model)\label{fig:teaser-b}]{
\includegraphics[width=0.28\linewidth]{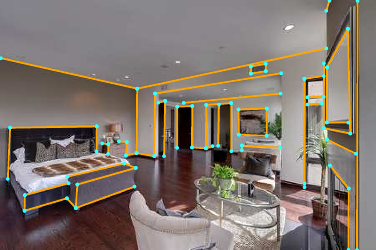}
\includegraphics[width=0.28\linewidth]{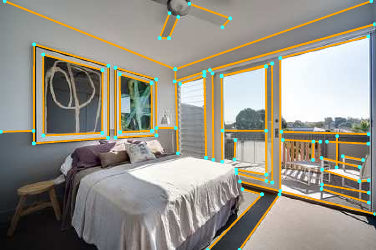}
\includegraphics[width=0.28\linewidth]{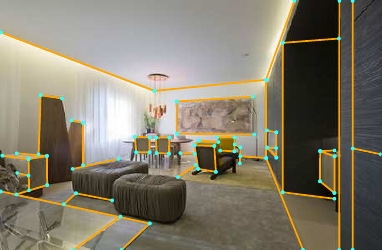}
}\vspace{-3mm}

\subfigure[HAWPv3 on the Wireframe Dataset (SSL Model)]{
\includegraphics[width=0.28\linewidth]{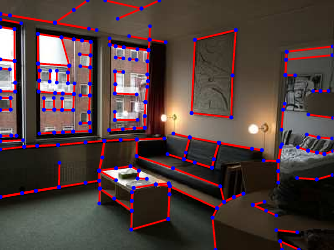}
\includegraphics[width=0.28\linewidth]{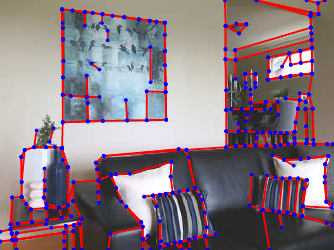}
\includegraphics[width=0.28\linewidth]{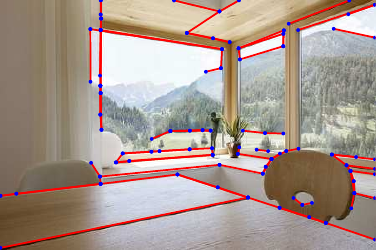}
}\vspace{-3mm}

\subfigure[HAWPv3 on the BSDS Dataset (SSL Model)]{
\includegraphics[width=0.28\linewidth]{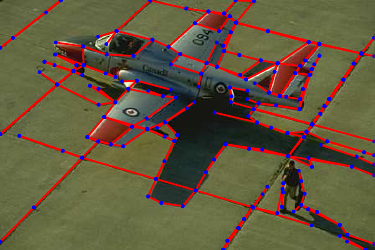}
\includegraphics[width=0.28\linewidth]{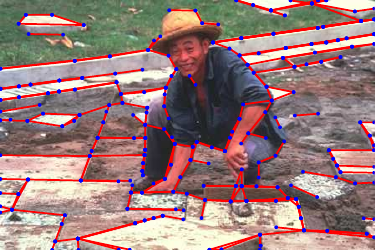}
\includegraphics[width=0.28\linewidth]{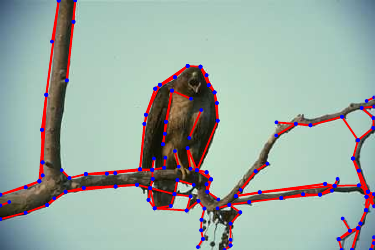}
}\vspace{-3mm}

\subfigure[HAWPv3 on the AICrowd Dataset (SSL Model)]{
\includegraphics[width=0.28\linewidth]{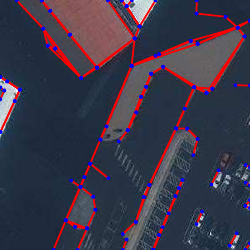}
\includegraphics[width=0.28\linewidth]{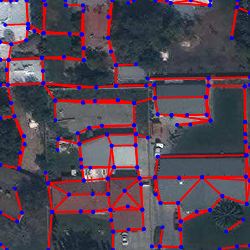}
\includegraphics[width=0.28\linewidth]{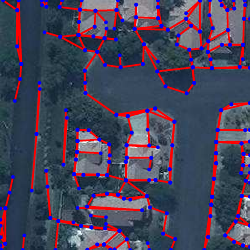}
}

\vspace{-2mm}
\caption{
The proposed HAWP models excel in wireframe structure perception using both fully-supervised learning (FSL) and self-supervised learning (SSL). HAWPv1~\cite{HAWP} and improved HAWPv2 are FSL models trained with human-annotated wireframes, primarily in indoor images. HAWPv3, an SSL model built on HAWPv2, enables wireframe parsing in out-of-distribution images such as those from the BSDS-500 dataset~\cite{bsds} and AICrowd dataset~\cite{AICrowdDataset}, without requiring labeled wireframes.
}
\label{fig:teaser}
\end{figure}

In this paper, our focus lies on modeling, learning, and parsing wireframes~\cite{wireframe-dataset} from images, which represent a parsimonious form of the primal sketch. As depicted in Fig.~\ref{fig:teaser}, wireframes capture line segments and their associated endpoints (primarily junctions) in images, enabling vectorized representations of the underlying boundary structures of objects and generic regions (stuff). Despite line segments being simple geometric patterns/symbols by definition, effectively modeling and computing them from images presents an exceedingly challenging problem due to the inherent uncertainty and ambiguity in grounding line segments to image pixels (referred to as the symbol-to-signal gap). 

We present a learning-based approach for wireframe parsing (Fig.~\ref{fig:hawp}), aiming to bridge the gap between pixels (2D signals) and symbols (line segments and junctions) progressively. Our method involves three key steps: (1) learning a novel Holistic Attraction (HAT) field representation that characterizes the geometry of line segments, including their endpoints, by incorporating both edge and non-edge pixels; (2) binding densely predicted line segments to a reduced set of junction/endpoint proposals, eliminating the need for complex non-maximum suppression (NMS); and (3) achieving wireframe parsing through a proposal verification module. This paper extends our previous work~\cite{HAWP} (HAWPv1, published in CVPR'20) with significant modifications in two aspects. Firstly, we address the effective and robust learning of the proposed HAT fields in the fully supervised learning setting, resulting in improved HAWPv2. Secondly, we broaden the applicability of our HAT fields through self-supervised learning, leading to HAWPv3 as an extension of HAWPv2 for wireframe parsing in diverse scenarios. Notably, our ablation studies in Appx.~D demonstrate that HAWPv1 does not support self-supervised learning well. We summarize and discuss the modifications and development path of the HAWP models as follows.

Our proposed HAT field representation stands out from other methods in the literature due to two novel aspects. Firstly, it incorporates a conceptually simple yet expressive line-segment-to-attraction-region lifting, which aligns with the population coding principle~\cite{averbeck2006neural} observed in primate visual systems and naturally incorporates visual context awareness. Secondly, it features a rigorously formulated closed-form differentiable HAT field parameterization (see Sec.~\ref{sec:representation} for detailed formulations), promoting representational parsimony and encouraging inferential consistency among the population, i.e., all (foreground) pixels within an attraction region. These aspects contribute to the stability and efficiency of learning HAWP. In fully supervised learning settings, follow-up studies on wireframe parsing~\cite{FClip,LETR,TP-LSD} without HAT fields typically require hundreds of training epochs to achieve comparable performances to our HAWPv1~\cite{HAWP}, which is trained in just 30 epochs. Intriguingly, the combination of these two aspects facilitates a significantly more efficient self-supervised learning paradigm.

In the development of HAWPv2, we explore novel aspects within the fully supervised learning paradigm. We investigate the closed-form property of the HAT field representation and unveil a simple yet effective differentiable loss function, which penalizes the endpoint fitting error in 2D Euclidean space for densely predicted line segments, thereby reducing invalid proposals. To address the densely predicted line segment proposals and leverage complementary endpoint/junction information, we propose a method for binding line segment proposals and endpoint/junction proposals. This approach significantly reduces the number of joint proposals by incorporating geometry co-occurrence, serving as an effective and efficient replacement for non-maximum suppression (NMS) of line segment pairs. Furthermore, we introduce a novel and lightweight verification module called {\em endpoint-decoupled LOIAlign} (EPD LOIAlgin). This module captures geometry-aware discriminative features for verification in a lightweight design. Taking advantage of HAT fields, which produce high-quality proposals together with informative point-line co-occurrence patterns by the endpoint predictions from heatmaps, the hand-crafted designs used in L-CNN~\cite{L-CNN} and HAWPv1 are {\bf no longer necessary to train the verification module}.

Obtaining ground-truth wireframes and other structural annotations for supervised learning is a time-consuming, costly, and often biased process. As a result, fully supervised wireframe parsers often struggle to produce satisfactory parsing results for images that differ significantly from the limited training data, leading to out-of-distribution failures. Motivated by this and aiming for more generalized wireframe parsers, we introduce HAWPv3, empowered by the expressive HAT field representation and inspired by the successes of SuperPoint~\cite{SuperPoint} for keypoint detection and SOLD$^2$~\cite{SOLD2} for self-supervised wireframe parsing.
To overcome the limitations of fully supervised learning, we adopt a simplified homographic adaptation pipeline from the self-supervised learning approach of SOLD$^2$. Using this approach, HAWPv3 achieves remarkable efficiency and effectiveness {\bf, requiring approximately 10 times less synthetic pretraining overall} and can be {\bf trained within 24 GPU hours on a single-GPU workstation}. In particular, it significantly increases the crucial metric of structural repeatability scores for line segments, improving from 61.6\% to 75.1\% in the wireframe dataset~\cite{wireframe-dataset}, and from 62.9\% to 71.1\% in the YorkUrban dataset~\cite{Denis2008}, respectively. These results underscore the expressive power of the HAT field representation and the elegantly designed wireframe parsing workflow. Fig.~\ref{fig:teaser} (c)-(e) provide illustrative examples of the impressive HAWPv3 results in three diverse datasets. 

In summary, this paper makes three key contributions to the field of wireframe parsing:

\begin{itemize}[leftmargin=*]
    \item The proposed line-segment-to-attraction-region lifting provides a new paradigm for learning expressive HAT fields for line segment representation, thanks to the richness of learnable information at the regional level. It shows significantly better effectiveness and expressiveness in both fully-supervised learning and self-supervised learning.  

    \item The proposed wireframe parsing frameworks (HAWPv2 and HAWPv3) are elegantly designed and well-cooked. The proposed line segment binding module and the proposed end-point-decoupled LOIAlign facilitate built-in robustness and geometry awareness in wireframe parsing.  

    \item In experiments, we demonstrate the superiority of our proposed HAWPv2 in supervised learning and our HAWPv3 in self-supervised learning with state-of-the-art performance obtained. Our HAWPv3 shows great potential for general wireframe parsing in images out of the training distributions. In addition, our codes and trained models are publicly available\footnote{https://github.com/cherubicXN/hawp}.
\end{itemize}

\paragraph*{Paper Organization.} The remainder of this paper is organized as follows. Sec.~\ref{sec:related-work} reviews the recent efforts in learning wireframe parsing and line segment detection and then summarizes our contributions. In Sec.~\ref{sec:representation}, we present the HAT field of line segments and discuss it with alternative representations. Sec.~\ref{sec:framework} and Sec.~\ref{sec:models} describe the framework of HAWP and the details of learning HAWPv2 and HAWPv3 models, respectively. In the experiments, we evaluate the proposed HAWPv2 model in Sec.~\ref{sec:exp-fsl} and HAWPv3 model in Sec.~\ref{sec:ssl-exp}. Last but not least, we conclude this paper in Sec.~\ref{sec:conclusion}.

\section{Related Work}\label{sec:related-work}
As mentioned earlier, wireframe parsing is a relatively new concept that focuses on modeling and computing line segments and junctions in 2D images, providing a concise representation for the robust geometric understanding of the visual world. The roots of wireframe parsing can be traced back to the early days of computer vision, such as Larry Roberts' work on understanding the "Blocks World"~\cite{roberts1963machine,GuptaEH10}, as well as the primal sketch concept proposed by David Marr~\cite{MarrH80}. Over time, significant progress has been made in developing more expressive yet parsimonious representations and powerful yet efficient parsing algorithms to enhance the geometric understanding of images. Notable examples include the alignment-based LSD framework~\cite{LSD} with an a-contrario verification before the era of modern deep learning. In this section, we review recent advancements in learning-based wireframe parsing, which is the domain to which the proposed HAWP models belong.

\paragraph*{Inductive and Deductive Wireframe Parsing.}
Wireframe parsing computation involves two phases: proposal generation and verification. For endpoint/junction proposals in wireframe parsing, methods commonly use heatmap regression. There are two main categories of approaches for line segment modeling in proposal generation: inductive parsing, which leverages line segment biases (e.g., learned line heatmaps) and deductive parsing, which enumerates all pairs of detected endpoints for line segment proposals.

The Deep Wireframe Parsing method and Wireframe dataset~\cite{wireframe-dataset} introduced an inductive approach, connecting endpoints using learned line heatmaps for line segment proposals. Our previous work on regional attraction fields~\cite{AFM-CVPR,RegionalAttraction} also falls into the inductive category, extracting line segments from attraction fields using a heuristic squeezing module. Deductive approaches~\cite{PPGNet,L-CNN} bypass challenges by enabling end-to-end training. L-CNN~\cite{L-CNN} improves parsing with LOIPooling for proposal verification. Deductive approaches tackle class-imbalance issues caused by exhaustive enumeration, requiring careful sampling strategies during training.
Subsequently, both our preliminary HAWPv1~\cite{HAWP} and the LGNN method~\cite{LGNN} build on the success of LOIPooling in L-CNN~\cite{L-CNN} while addressing the inefficiency of line segment proposal generation. These approaches demonstrate that modeling the inductive biases of line segments in a more direct and appropriate manner can further enhance the accuracy and efficiency of wireframe parsing, all while retaining the benefits of end-to-end training. The LGNN method~\cite{LGNN} introduces the center-offset representation for line segments, which, as discussed in Sec.~\ref{sec:alternativefields}, may exhibit slower convergence during the learning process.

Compared with prior art, the proposed HAWPv2 has the core HAT field representation first proposed in our preliminary HAWPv1, and harnesses the best of both line segment and endpoint proposals (the co-occurrence modeling and the end-point-decoupled LOIAlign as briefly discussed in Sec.~\ref{sec:introduction}) in a systematic way. The proposed HAWPv3 further extends the horizon of how wireframes can be learned by developing an effective SSL paradigm.

\paragraph*{Line Segment Representations.}
As one of the most primitive geometric patterns/symbols, line segments are easy to describe and define mathematically, but have been shown to be extremely challenging to induce their conceptually simple inductive biases end-to-end in wireframe parsing. The key question is how to parameterize line segments in a differentiable way in the rasterized image lattice. There are three types of formulations that allow fully end-to-end training with different levels of effectiveness. 

\textit{The center-offset and query-to-endpoints representations.} 
Unlike the exhaustive enumeration in deductive approaches, alternative inductive representations based on center offset and Transformers~\cite{vaswani2017attention,carion2020end} have been explored for line segment detection and wireframe parsing. Inspired by the "objects as points" concept in object detection~\cite{CenterNets-Zhou}, line segments are parameterized by their center point, tangent angle, and Euclidean length in TP-LSD~\cite{TP-LSD}, F-Clip~\cite{FClip}, M-LSD~\cite{gu2021realtime}, and E-LSD~\cite{ELSD}. Meanwhile, LETR~\cite{LETR} transforms latent queries into the endpoints of line segments using attention mechanisms. These representations, focusing on points along the line segments, are sparse and lack explicit context awareness. Consequently, these methods often require extensive training epochs (e.g., 300 epochs in F-Clip and 800 epochs in LETR).

\textit{The HAT field representation.} It is first proposed in our preliminary HAWPv1~\cite{HAWP} before the two representations stated above. In contrast to them, the HAT field lifts line segments to non-overlapping attraction regions following our previous work on the regional attraction field representation~\cite{AFM-CVPR,RegionalAttraction}. It then forms context-aware population coding for a line segment using ``foreground" points in the attraction region with points on the line segment excluded.  As a result, our HAWP models often use vanilla convolution operations and need only 30 epochs in training with state-of-the-art performance obtained in testing.

\paragraph*{Self-Supervised Learning of Wireframe Parsing.}
Seeking proper pre-text tasks~\cite{he2020momentum, he2022masked} or self-consistency prediction/matching is one of the keys to self-supervised learning of modern deep vision models. In terms of the wireframe parsing and the related geometric tasks, pipelines that explore the simulation-to-reality workflow has also been studied. 
By utilizing the simulation-to-reality workflow, the SuperPoint~\cite{SuperPoint} presented the first self-supervised learning method for interest point detection and description with the proposed Homography Adaptation approach. Inspired by the SuperPoint, SOLD$^2$~\cite{SOLD2} presents the first self-supervised learning method for line detection using the deductive wireframe parsing framework similar to L-CNN~\cite{L-CNN}. 

SOLD$^2$ requires a heavy synthetic pretraining using 160k synthetic data. Due to its deductive nature, it also suffers from the severe class-imbalance issue in the proposal verification, but cannot use the carefully-designed sampling strategy proposed in L-CNN in SSL. It thus resorts to the edge maps as the primal cues for verifying proposals in training with real unlabeled images, and achieves better performance in terms of detection repeatability across viewpoints, thus facilitating the learned detectors in downstream tasks such as SLAM and SfM. Our proposed HAWPv3 adopts the overall SSL pipeline used by SOLD$^2$. The expressiveness and effectiveness of the core HAT field representation result in a much more efficient SSL (\eg, 10x less synthetic pretraining cost) while being more accurate in terms of detection repeatability.

\section{HAT Fields of Line Segments}\label{sec:representation}
In this section, we present details of the proposed end-to-end trainable HAT field representation that introduces a unified parsimonious closed-form 4D geometric vector space to encode line segments in a wireframe. We also discuss two alternative field representation schema for line segments and give a high-level explanation of why the proposed HAT field representation is better. %

\subsection{Notations}

A wireframe is composed of a set of line segments. These line segments are represented using real coordinates (i.e., sub-pixels) within the continuous image domain $D \subset \mathbb{R}^2$. We also utilize the representation of line segments over the discrete image lattice $\Lambda \subset D$, which consists of integer pixel coordinates (for example, $512\times 512$ pixels) in rasterization.
We denote a line segment as $\lseg = (\mathbf{x}1, \mathbf{x}2) \in D \times D$. Here, the symbol $\ddot{~}$ is used to highlight the two endpoints of a line segment. The line that passes through the line segment $\ddot{\bm l}$ in $\mathbb{R}^2$ is denoted by $\line = (\mathbf{a}_{\lseg}, b_{\lseg})$. In this case, $\mathbf{a}_{\lseg} \in \mathbb{R}^2$ and $b_{\lseg}\in\mathbb{R}$ represent the coefficients (slope and intercept) of the line equation, which are uniquely determined by $(\mathbf{x}_1, \mathbf{x}_2)$.

\subsection{The Attraction Region of a Line Segment}
Utilizing the method from our previous work~\cite{AFM-CVPR,RegionalAttraction}, we introduce an attraction region for each line segment to counter its inherent locality. As depicted by the gray region in Fig.~\ref{fig:HAF} (a), each point $\mathbf{p}\in \Lambda$ is assigned to the line segment $\lseg$ to which it is closest. Distance is computed by projecting $\mathbf{p}$ onto the line $\line$ of $\lseg$ to get $\mathbf{p}'\in D$. If $\mathbf{p}'$ is not on $\lseg$, it's reassigned to the nearest endpoint. The distance between $\mathbf{p}$ and $\lseg$ is the Euclidean distance between $\mathbf{p}$ and $\mathbf{p}'$. Attraction regions of different line segments are disjoint, enabling regional representation of line segments and facilitating more effective learning.

Once the attraction region of a line segment is computed, any pixel within this region can "recover" the line segment, given a suitable encoding scheme. This approach effectively counters the inherent locality and corresponding ambiguity of line segments. However, the question arises: \textbf{What is the suitable encoding scheme that enables this representational effectiveness?} In the following sections, we will first introduce the proposed parsimonious closed-form 4D geometric vector field. Subsequently, we will discuss alternative representations that are less effective in experimental comparisons, as shown in Fig.~\ref{fig:recall-cmp-fields} and Sec.~\ref{sec:exp-fsl}.

\subsection{The HAT Field of a Wireframe}
\begin{figure}
    \centering
    \includegraphics[width=0.99\linewidth]{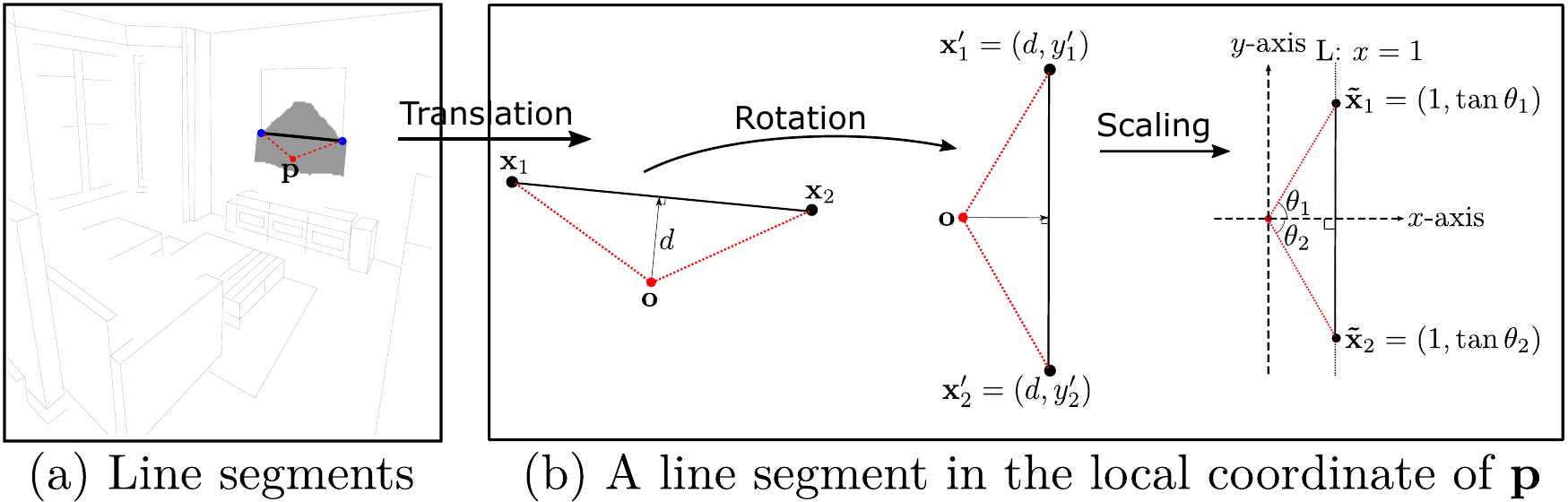}
    \caption{Illustration of constructing the proposed HAT field representation for wireframes. (a) displays an example of wireframes, highlighting one line segment in black with its two endpoints in blue, and the corresponding attraction region shaded in solid gray. (b) demonstrates the derivation of the 4D geometric vector representation for a "foreground" point/pixel $\mathbf{p}$ within the attraction region. See text for details.
    }
    \label{fig:HAF}
\end{figure}

\begin{figure*}
    \centering
    \subfigure[\scriptsize A toy example\label{fig:toy-a}]{
    \includegraphics[width=0.22\linewidth]{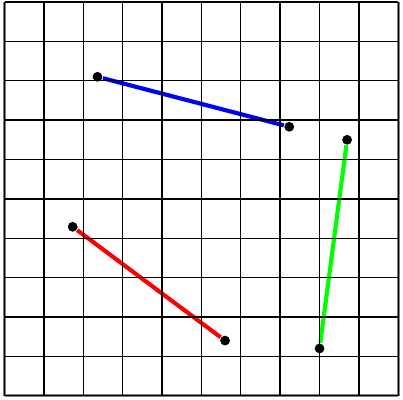}
    }
    \subfigure[\scriptsize Center-offset Vectors\label{fig:toy-center}]{
    \includegraphics[width=0.22\linewidth]{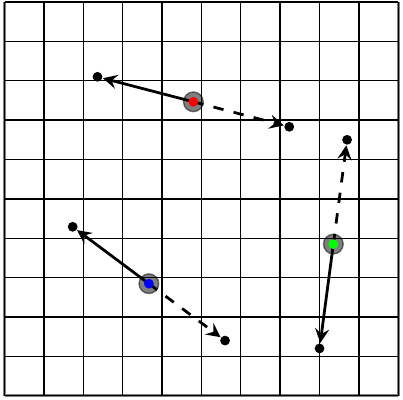}
    }
    \subfigure[\scriptsize 4D Attraction Field\label{fig:toy-afm}]{
    \includegraphics[width=0.22\linewidth]{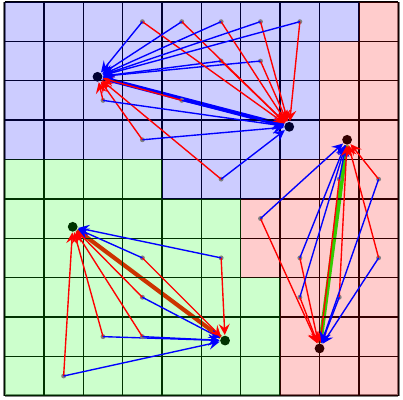}
    }
    \subfigure[Holistic Attraction Field\label{fig:toy-hat}]{
    \includegraphics[width=0.22\linewidth]{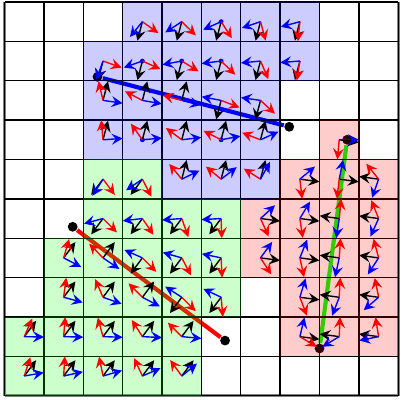}
    }
    \caption{
    Illustrative comparisons of different representations of a toy example \subref{fig:toy-a}. In \subref{fig:toy-center}, the center offset representation parameterizes three line segments by three center points and the corresponding offset vectors. In \subref{fig:toy-afm}, the 4D attraction field uses all pixels to explicitly characterize line segments as two offset vectors. In \subref{fig:toy-hat}, three angles $\theta$ (in black arrows), $\theta_1$ (in red arrows), and $\theta_2$ (in blue arrows) characterize the two endpoints of the corresponding line segment in a ``normalized" coordinate frame and transform the ``normalized" coordinate representation to the image coordinate by the distance component. The distance component is hidden to clearly show the angle components.
    }
    \vspace{-5mm}
    \label{fig:toy}
\end{figure*}

The HAT field of a wireframe can be defined either in the same image domain $D$ or in a downsampled one to comply with the design of ConvNets, denoted by $\mathcal{A}$. For each point $\mathbf{p}\in \Lambda$, $\mathcal{A}(\mathbf{p})$ is a 4D geometric vector derived as follows.

As shown in the right columns in Fig.~\ref{fig:HAF},  for a line segment $\ddot{l}$, our derivation undergoes a simple affine transformation for each distant point $\mathbf{p}$ in its attraction region. Let $d$ be the distance between $\mathbf{p}$ and $\ddot{l}$, \ie, $d=|\mathbf{a}^{\top}_{\ddot{l}}\cdot \mathbf{p}' + b_{\ddot{l}}|>0$. We have,  
\begin{enumerate}%
    \item [i)] \textit{Translation}: The point $\mathbf{p}$ is then used as the new coordinate origin. 
    \item [ii)] \textit{Rotation}: The line segment is then aligned with the vertical $y$-axis with the end-point $\mathbf{x}_1$ on the top and the point $\mathbf{p}$ (the new origin) to the left. The rotation angle is denoted by $\theta\in [-\pi, \pi)$.
    \item [iii)] \textit{Scaling}: The distance $d$ is used as the unit length to normalize the $x$- / $y$-axis in the new coordinate system.  
\end{enumerate}

In the new coordinate system after the affine transformation, let $\theta_1 \in (0,\frac{\pi}{2})$ and $\theta_2 \in (-\frac{\pi}{2},0]$ be the two angles as illustrated in Fig.~\ref{fig:HAF}. So, a point $\mathbf{p}$ in the attraction region of a line segment $\ddot{l}$ is reparameterized as, 
\begin{equation}
    \mathbf{p}(\lseg) = (d, \theta, \theta_1, \theta_2), \label{eq:HAF}
\end{equation}
which can exactly recover the line segment in a closed form in the ideal case using, 
\begin{equation}%
    \lseg = d\cdot \begin{pmatrix}\cos \theta & -\sin\theta \\ \sin\theta & \cos \theta \end{pmatrix}\begin{pmatrix}1 & 1 \\ \tan\theta_1 & \tan\theta_2 \end{pmatrix} + (\mathbf{p}^{\top},\mathbf{p}^{\top}), \label{eq:lineseg}
\end{equation}
where an estimated 4D vector $(\hat{d}, \hat{\theta}, \hat{\theta_1}, \hat{\theta_2})$ will generate a line segment proposal. 

\textbf{Differentiability of the Proposed Line Segment Representation.} Based on Eqn.\eqref{eq:lineseg}, a significant merit of the proposed holistic attraction field representation is its differentiability at every point in the attraction region. This 4D geometric parameterization of line segments leads to a new loss function, as proposed in Eqn.\eqref{eq:linseg_loss}, for training.

\textbf{Foreground and Background Mask of a Holistic Attraction Field.} Certain points (pixels), such as those on any line segments, should not be reparameterized to prevent degradation and are thus classified as the "background". Additionally, we create "shrunk" attraction regions for line segments by excluding points whose distances exceed a predefined threshold $\tau_d$ (defined later). This approach allows the model to concentrate more on the immediate surrounding context. We denote the mask by $\mathbf{M}(\mathbf{p})=1$ for the foreground and $\mathbf{M}(\mathbf{p})=0$ for the background. Background points not attracted by any line segment, as per our specification, are encoded using a 4-D null vector.

So, computing line segment proposals for a wireframe is posed as a map-to-map (\ie, image-to-attraction-field) regression problem, which enables us to exploit many encoder-decoder neural networks in learning.

\subsection{Alternative Attraction Field Representations}\label{sec:alternativefields}
For comparisons, we discuss two alternative encoding schema of vector field representations for line segments as shown in Fig.~\ref{fig:toy} using a toy example.
\begin{figure}
    \centering
    \includegraphics[width=0.9\linewidth]{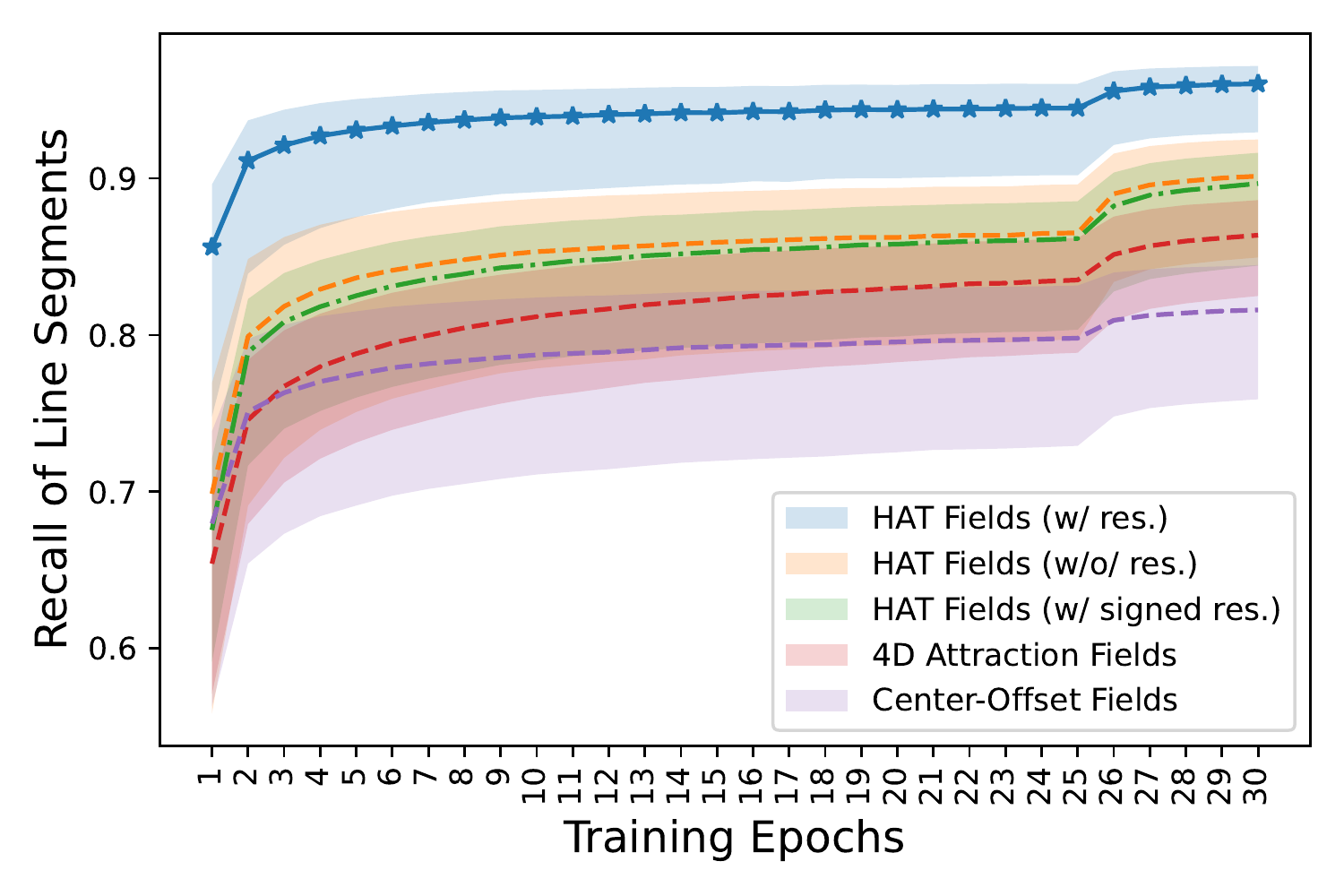}
    \vspace{-4mm}
    \caption{Comparison of training convergence in terms of line segment recall rates among different field representations. Representations compared include our proposed HAT fields (with and without unsigned residual learning) and {an alternative to signed residual learning}, the 4D attraction fields, and the center-offset fields. Within each colored region, we compute the recall rates at various training epochs using different thresholds $\vartheta \in [5,15]$ as defined in Eq.~\eqref{eq:sap-dis}, with the results for $\vartheta=10$ marked in line patterns.}
    \vspace{-5mm}
    \label{fig:recall-cmp-fields}
\end{figure}
\paragraph*{The Regional Attraction Field Representation.}
It is a straightforward extension of our previous regional attraction work~\cite{AFM-CVPR,RegionalAttraction}. Similarly, it also utilizes the attraction region lifting for line segments. As illustrated in Fig.~\ref{fig:HAF}(a) and Fig.~\ref{fig:toy-afm}, consider a distant pixel point $\mathbf{p}$ outside a line segment $\lseg$, with the projection point $\mathbf{p}'$ being on the line segment. The conventional regional attraction method reparameterizes $\mathbf{p}$ as $\mathbf{p}-\mathbf{p}'$, i.e., the displacement vector in the image domain. This results in a 2-D vector field for a wireframe by reparameterizing all the pixel points in the image lattice $\Lambda$. If we use the two displacement vectors between $\mathbf{p}$ and the two endpoints of the line segment, we can reparameterize $\mathbf{p}$ by its 4D displacement vector, which can completely determine the line segment (i.e., an exact dual representation). By including the displacement vector $\mathbf{p}-\mathbf{p}'$, we can obtain a 4D/6D field representation, depending on whether the perpendicular vectors are encoded.

In the process of developing the proposed HAT field representation, we initially used the 4D/6D regional attraction field representation. However, we found that these representations could not be accurately and reliably learned during training with deep neural networks (DNNs). Despite the 4D/6D regional attraction field's capability to capture all the necessary information for recovering line segments in a closed-form way, it contains large structural variations that pose challenges for learning. Our observations suggest that DNNs are not sufficiently effective in learning and predicting this type of information.

In contrast, the proposed HAT field representation incorporates both distance and angles, defined in a line-segment-aware local coordinate system (via the aforementioned affine transformation), making it holistic. The combination of distance and angle information helps address the learnability issue associated with solely fitting distance or displacement. It's important to note that even with the holistic attraction field representation, we still need to employ a residual learning method to predict the distance component. In Fig.~\ref{fig:recall-cmp-fields}, we compare the different strategies of learning unsigned and the signed distance residuals, showing that the signed distance residual learning will lead to a lower recall rate even than the model without residual learning.

\paragraph*{The Center-Offset Field Representation.} 
In recent literature, the center-offset field representation has been extensively used to represent annotated line segment maps (~\cite{TP-LSD,FClip}, to cite a few). Given a line segment $\lseg = (\mathbf{x}_1, \mathbf{x}_2)$, it is encoded by the center point $\mathbf{x}_c = (\mathbf{x}_1+\mathbf{x}_2)/2$, the tangent angle $\alpha$ (or its $\cos$ counterpart) and the length $l = \left\|\mathbf{x}_1 -\mathbf{x}_2\right\|_2$. Although the center-offset field representation enjoys a simple formulation, it is a sparse field and lacks the richness contributed by lifting line segments to attraction regions in our HAT field representation. Consequently,  learning the center-offset field representation often suffers from slow convergence. For example, the best-performing detector, F-Clip~\cite{FClip}, entails a very long learning schedule (of 300 epochs) to achieve promising results to overcome the difficulty of predicting the length component accurately and reliably.

We present the comparisons between our HAT field representation and the two alternatives in Fig.~\ref{fig:recall-cmp-fields}, which shows that our HAT field representation enables a much faster learning convergence in terms of the recall of line segments (see Sec.~\ref{sec:fsl-setting} and  Sec.~\ref{sec:exp-fsl} for experimental settings). But, why? We provide a high-level explanation as follows.

\paragrapha*{Why is the HAT field an ``Attractive" Representation?} 

{
In addition to the computational merit of being differentiable (Eqn.~\ref{eq:lineseg}), the proposed holistic attraction field also has a strong intuition that has not been fully exploited by the alternatives. As a toy example in Fig.~\ref{fig:toy-a} that has 3 line segments defined on the image grid, the center-offset representation (and its variant) has to face two challenging cases to be solved: (1) {how to accurately detect the center locations} and (2) {how to cope with the large variation for the small-length and large-length line segments that are presented in the same image to accurately regress the length of line segments}. As a result, the methods built on the center-offset representations usually have long training schedules with hundreds of epochs. To further improve the learning ability, some hand-crafted neural modules based on \cite{DCNv1,deeplab} were studied~\cite{TP-LSD,ELSD,FClip} for feature aggregation.

Different from center-offset representations, which are strictly defined on center pixels, the regional representations in Fig.~\ref{fig:toy-afm} and Fig.~\ref{fig:toy-hat} have a more relaxed definition by incorporating non-edge pixels. With the involvement of more pixels, there is no need to precisely localize the foreground pixels during learning; instead, the focus can solely be on the regression of the fields. However, the 4D attraction field in Fig.~\ref{fig:toy-afm}, which utilizes long-range vectors to depict line segments, is susceptible to large structural variations during the learning process.
By contrast, our HAT field representation normalizes the two long-range vectors using three angles (or unit vectors) in the local coordinate system, and incorporates the distance to transform the local line segments into the image coordinate system. This type of "representation normalization" eliminate many nuisance factors in the data, thereby facilitating more effective learning. Moreover, the joint encoding that exploits displacement distance and angle effectively decouples the attraction field with respect to complementary spanning dimensions.

}

\section{The Proposed HAWP Framework}\label{sec:framework}
\begin{figure*}
    \centering

    \begin{tabular}{c|c}
         \subfigure[\scriptsize Proposal Generation\label{subfig:featurization}]{
    \includegraphics[height=.2\linewidth]{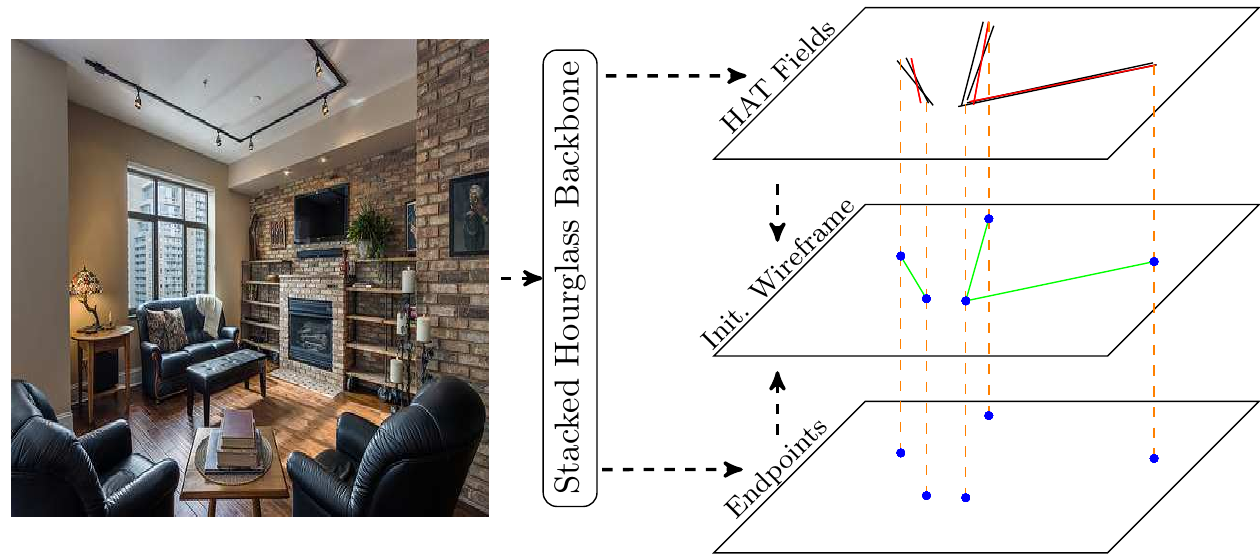}
    } & \subfigure[\scriptsize EndPoint-Decoupled LOIAlign\label{subfig:verification}]{
    \includegraphics[height=.23\linewidth]{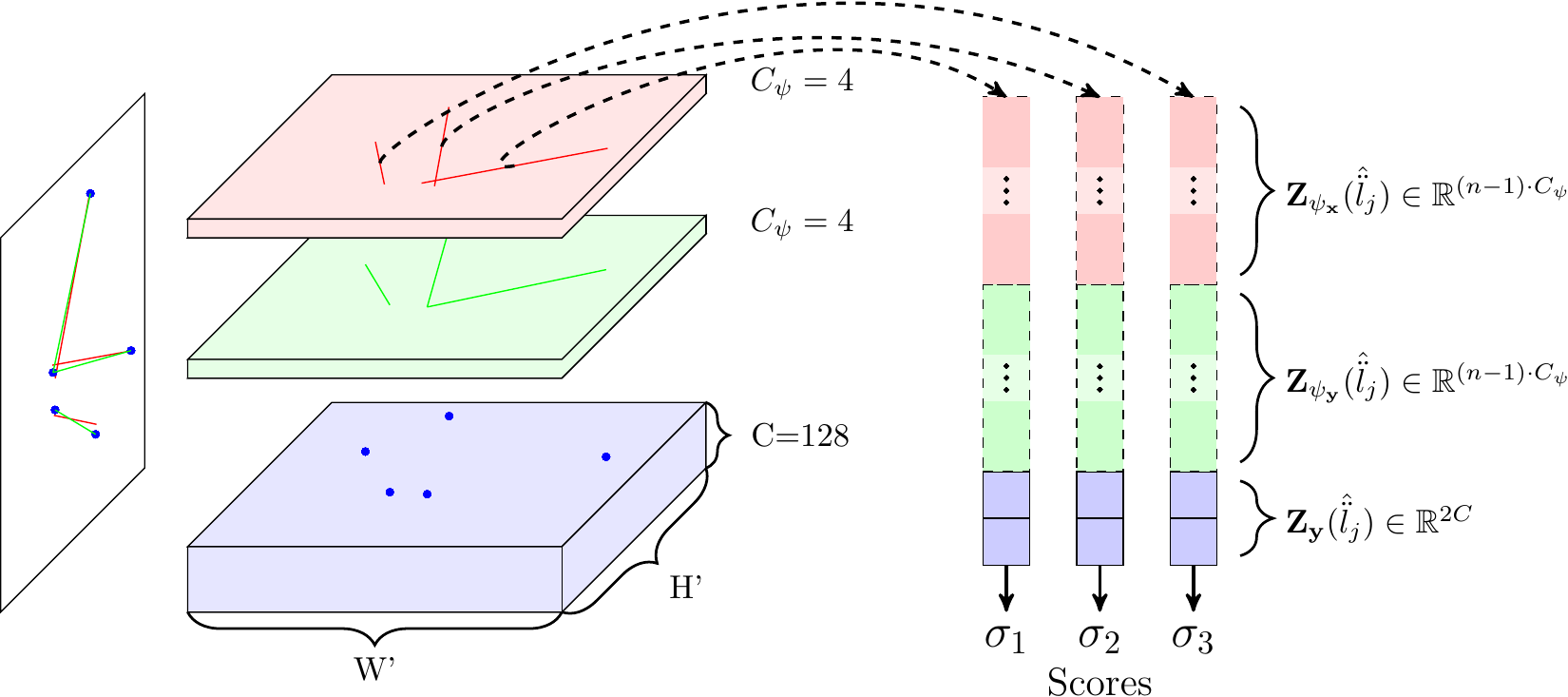}
    }
    \end{tabular}

    \caption{
    The overall architecture of the proposed HAWP framework. In \subref{subfig:featurization}, the HAT Fields and Junctions are predicted from the shared backbone network to yield initial wireframe proposals by binding the line segments (from the predicted HAT field) and junctions (from the predicted heatmap) together into the green line segments; In \subref{subfig:verification}, we present an EPD LOIAlign module to extract the endpoint features from the $C$-dim feature map, and the HAT-augmented line segment features from two thin feature maps with the $C_{\psi}$-dim for the HAT line segment (in red) and the junction-refined line segment (in green). With the HAT-sourced line segment as an augmentation, the co-occurrence patterns between the HAT fields and the junction predictions are captured for better verification, facilitating the learning of HAWPv2 and HAWPv3 models. 
    }
    \label{fig:hawp}
\end{figure*}

In this section, we present details of the proposed HAWP framework that is built on the HAT field representation of a wireframe.  As illustrated in Fig.~\ref{fig:hawp}, the proposed HAWP consists of three components: line segment and  proposal generation (Sec.~\ref{sec:learning-hat} and Sec.~\ref{sec:learning-juncs}), line segment and  binding (Sec.~\ref{sec:binding}), and endpoint-decoupled line-of-interest verification (Sec~\ref{sec:loi-pooling}).

\subsection{Notations}
Let $I$ represent an image defined on the image lattice $\Lambda$ with dimensions $H\times W$ pixels (e.g., $512\times 512$). For the wireframe associated with image $I$, we define $\ddot{\mathbf{L}}$ as the set of annotated line segments in $I$, and $\dot{\mathbf{J}}$ as the set of unique endpoints belonging to all line segments in $\ddot{\mathbf{L}}$. It is worth noting that many of these endpoints are junction points formed by multiple line segments.

We denote $f_b(\cdot;\Omega_b)$ as the deep neural network feature backbone with parameters $\Omega_b$. Given an input image $I$, the feature backbone produces an output feature map $F=f_b(I)$ with dimension $C$, height $H_s = \frac{H}{s}$, and width $W_s = \frac{W}{s}$. The output resolution depends on the overall stride $s$ of the backbone, and the resulting lattice is $\Lambda'$, a sub-sampled version of the original lattice $\Lambda$.
On top of the feature map $F$, we employ lightweight head sub-networks to regress both the 4D HAT field for line segments and the endpoint heatmap. 
{\em Detailed network architectures can be found in Appx.~\ref{appx:networks}}. In this section, we refer to them as general mapping functions in defining our HAWP.

The predicted HAT field is computed at the resolution of $H_s \times W_s$, we prepare the ground-truth field maps at the same resolution for computing the loss of the predicted holistic attraction field in training accordingly. Since a wireframe is a vectorized representation, the mapping from the original image lattice $\Lambda$ to the (sub-sampled) lattice $\Lambda'$ is straightforward. Without loss of generality, we directly use the lattice $\Lambda'$ when referring to line segments and junction points in the formulations hereafter.

\subsection{Learning the HAT Field of Line Segments}\label{sec:learning-hat}

We use two separate head sub-networks in learning the distance and the three angles in our proposed 4D holistic attraction field $\mathcal{A}$. 
\begin{align}
    \text{The distance map:} \quad \mathcal{A}_d &= f_d(F;\Omega_d) \in \mathbb{R}^{1\times \Lambda'}, \label{eq:distance} \\
    \text{The angle field:} \quad \mathcal{A}_a & = f_a(F;\Omega_a) \in \mathbb{R}^{3\times \Lambda'}. \label{eq:angle}
\end{align}

Considering an annotated line segment $\ddot{l} \in \ddot{\mathbf{L}}$, a foreground point $\mathbf{p}'\in \Lambda'$ within the attraction region of $\ddot{l}$ is reparameterized as $\mathbf{p}'(\ddot{l})=(d, \theta, \theta_1, \theta_2)$ using Eq.~\eqref{eq:HAF}. Here, $\mathbf{p}'=\lfloor \mathbf{p}/s\rfloor \in \Lambda'$ is the mapped point from $\mathbf{p}\in \Lambda$. The ground-truth distance map and angle field are denoted as $\mathcal{A}_d^{gt}$ and $\mathcal{A}_a^{gt}$, respectively.
To facilitate neural network training, we normalize the distance map by $\min(\max(d/\tau_d,0),1)$, where $\tau_d$ controls the foreground pixels. The angles are normalized as $(\frac{\theta}{2\pi}+\frac{1}{2}, \frac{\theta_1}{\pi/2}, \frac{\theta_2}{\pi/2}+1)$. This normalization ensures that the predicted values of the HAT field fall within the range of $[0,1]$, and they can be unnormalized to obtain line segment proposals.

\begin{figure}
    \centering
    \resizebox{0.9\linewidth}{!}{
    \begin{tabular}{ccc}
         \includegraphics[width=0.3\linewidth]{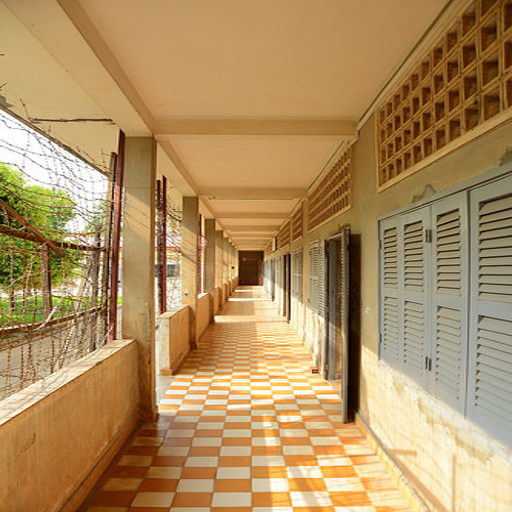} &
         \includegraphics[width=0.3\linewidth]{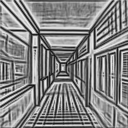} &
         \includegraphics[width=0.3\linewidth]{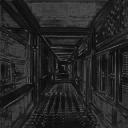}
         \\
         \includegraphics[width=0.3\linewidth]{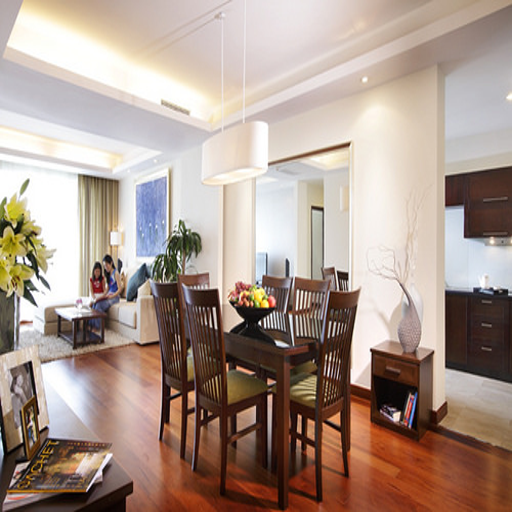} & 
         \includegraphics[width=0.3\linewidth]{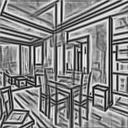} & 
         \includegraphics[width=0.3\linewidth]{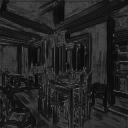}
         \\
         Inputs & Dis. Pred. & Res. Pred.
    \end{tabular}
    }
    \caption{Some illustrative examples for the learning of distance and residual maps. For each input image, the predicted distance (Dis. Pred.) and the absolute distance residuals (Res. Pred.) are displayed. The darker color is corresponding to the smaller values.}
    \label{fig:residual-learning}
\end{figure}

\paragraph*{Residual Learning of the Distance Map.}
As aforementioned, we observe that the distance map is more difficult to learn than the angle fields in our experiments. To address this, we propose a residual learning method by introducing another light-weight sub-network to predict the distance residual, 
\begin{equation}
    \mathcal{A}_{\Delta d} = f_{\Delta d}(F;\Omega_{\Delta d}) \in \mathbb{R}^{1\times \Lambda'}. \label{eq:dresidual}
\end{equation}
The ground-truth of the distance residual map is computed on the fly,  $\mathcal{A}^{gt}_{\Delta d}(\mathbf{p}') = |\mathcal{A}^{gt}_d(\mathbf{p}') - \mathcal{A}_d(\mathbf{p}')| $.  Fig.~\ref{fig:residual-learning} shows some examples of the learned distance maps and residual maps. 

{
Note that we learn the unsigned residual instead of the signed one, which is mainly due to that the distance field and its residual take the same feature map as input to avoid unnecessary increase of model complexity. Furthermore, the learning of unsigned residual can be viewed as a kind of uncertainty estimation for the distance prediction, thus facilitating the learning of HAT fields as shown in our experiments. Given by this, we enumerate their signs and modulate the scales to get a set $K = \{-k,\ldots,0,\ldots,k\}$ ($k$ is a hyperparameter, to be specified in experiments), and eventually yield a set of rectified distance maps by, 
}
\begin{equation}\label{eq:decoding-res}
\mathcal{A}^{(i)}_d(\mathbf{p}') = \mathcal{A}_d(\mathbf{p}') + i\cdot \mathcal{A}_{\Delta d}(\mathbf{p}'), ~~\forall i \in K. 
\end{equation}
With the rectified distance maps, $2k+1$ line segment proposals will be generated at each foreground point and refined by the line-of-interest verification step.  

\paragraph*{Loss Functions.}
We use the $\ell_1$ loss function between the ground-truth maps $\mathcal{A}^{gt}_d$,  $\mathcal{A}^{gt}_a$, $\mathcal{A}^{gt}_{\Delta d}$ and the corresponding predicted maps $\mathcal{A}_d$,  $\mathcal{A}_a$, $\mathcal{A}_{\Delta d}$. The loss is computed across the foreground points only based on the mask $\mathbf{M}(\mathbf{p}')$. 

Due to the multiple line segment proposals generated by the residual learning of the distance map at each foreground point, we propose a loss function to directly minimize the endpoint error, $\mathcal{L}_{epe}$, between the line segment proposal and the ground one, thanks to the differentiability of our holistic attraction field (Eqn.~\ref{eq:lineseg}). {As the unsigned distance residual will generate $2k+1$ line proposals for each foreground pixel $\mathbf{p}'$, we encourage all the $2k+1$ proposals to be close to the line segment $\ddot{l}^{gt}(\mathbf{p}')$, and get the loss $\mathcal{L}_{epe}$ by}

\begin{equation}
    \mathcal{L}_{epe} = \sum_{i=-k}^{k}\sum_{\mathbf{p}'\in \Lambda'} \frac{\mathbf{M}(\mathbf{p}')}{\left\|\ddot{l}^{gt}(\mathbf{p}')\right\|} \ell_1(\ddot{l}^{(i)}(\mathbf{p}'), \ddot{l}^{gt}(\mathbf{p}')), \label{eq:linseg_loss}
\end{equation}
where $\ddot{l}^{(i)}(\mathbf{p}')$ is the $i$-th line segment proposal in the residual learning at the point $\mathbf{p}'$ and the $\ddot{l}^{gt}(\mathbf{p}')$ the ground-truth line segment.  $\left\| \cdot \right\|$ represents the length of a line segment. 

The total loss $\mathcal{L}_{\ddot{\mathbf{L}}}$ of learning line segments via the HAT field is simply the sum of the different $\ell_1$ loss terms (without trade-off tuning parameters). 

\begin{figure}
    \centering
    \fbox{\includegraphics[height=0.15\linewidth]{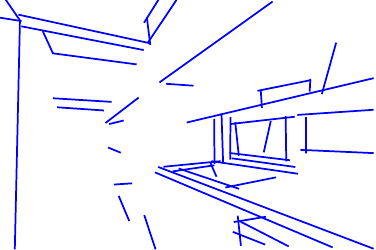}}
    \fbox{\includegraphics[height=0.15\linewidth]{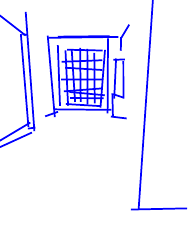}}
    \fbox{\includegraphics[height=0.15\linewidth]{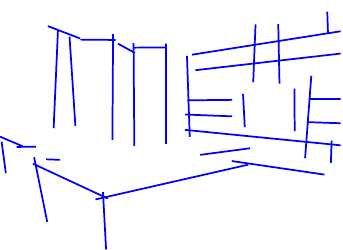}}
    \fbox{\includegraphics[height=0.15\linewidth]{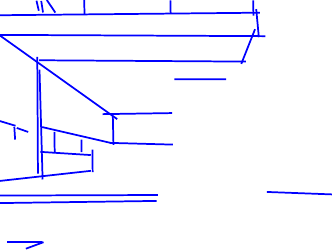}}
    \\\vspace{2mm}
    \fbox{\includegraphics[height=0.15\linewidth]{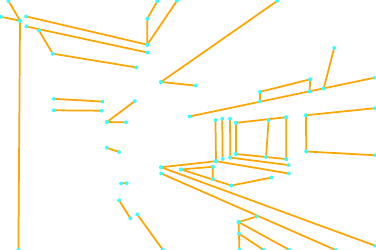}}
    \fbox{\includegraphics[height=0.15\linewidth]{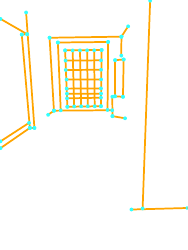}}
    \fbox{\includegraphics[height=0.15\linewidth]{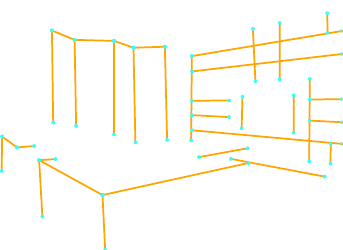}}
    \fbox{\includegraphics[height=0.15\linewidth]{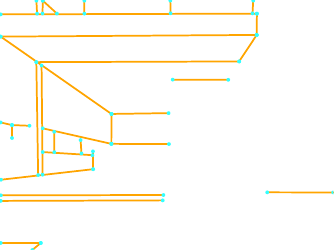}}
    \\
    \caption{Some illustrative examples for the learned line segments by HAT fields (top) and the corresponding wireframes (bottom) that are refined by the endpoint binding. Best viewed in magnification.}
    \label{fig:lineseg_proposals}
\end{figure}
Fig.~\ref{fig:lineseg_proposals} shows some examples of generated line segment proposals via learning the HAT field. We can see that the richness of our HAT field also leads to dense proposals of line segments that need to be refined for the final wireframe parsing. One straightforward way is to directly learn to score the line segment proposals followed by the Intersection-over-Union (IoU) based NMS, similar in spirit to object detection systems in~\cite{Faster-RCNN,CenterNets,CenterNets-Zhou}, however, it is not only time-consuming but also ill-defined (unlike that of bounding boxes). 
We address this issue by learning endpoints and then binding them with line segment proposals for pruning false positive line segment proposals.

\subsection{Learning the Heatmap-Offset Field of Endpoints}\label{sec:learning-juncs}
The heatmap-offset field representation is proposed in the L-CNN~\cite{L-CNN}. The heatmap regression is a widely used method in different types of keypoint detection (\eg, in human pose estimation).  The inclusion of an offset field aims to enhance the accuracy of sub-pixel localization for endpoints.
In this approach, the heatmaps for endpoint regression and the offset field are computed by two separate head sub-networks on top of the feature backbone output $F$:
\begin{align}
    \text{The endpoint heatmap:} \quad \dot{\mathbf{J}}_{pt} &= f_{pt}(F;\Omega_{pt}) \in \mathbb{R}^{1\times \Lambda'}, \label{eq:endpt} \\
    \text{The offset field:} \quad \dot{\mathbf{J}}_o & = f_o(F;\Omega_o) \in \mathbb{R}^{2\times \Lambda'}. \label{eq:offset}
\end{align}

The corresponding ground-truth maps are straightforwardly defined. Based on the set of annotated endpoints $\dot{\mathbf{J}}$ in an image, we use the ``hard" (binary) heatmap as the ground truth, 
\begin{equation}
    \dot{\mathbf{J}}_{pt}^{gt}(\mathbf{p}') = 1, \forall \mathbf{p}\in \dot{\mathbf{J}},
\end{equation}
where $\mathbf{p}'=\lfloor \mathbf{p}/s\rfloor \in \Lambda'$ and $\mathbf{p}\in \Lambda$. The associated offset is then defined by, 
\begin{equation}
    \dot{\mathbf{J}}_o^{gt}(\mathbf{p}') =  \mathbf{p}/s - \mathbf{p}'
\end{equation}

\paragraph*{Loss Functions.} 
We use the binary cross-entropy loss $\text{BCE}(\cdot,\cdot)$ for the endpoint heatmap regression, and the $\ell_1$ loss for the offset field,
\begin{align}
    \mathcal{L}_{pt} &= \text{BCE}(\dot{\mathbf{J}}_{pt}, \dot{\mathbf{J}}_{pt}^{gt}),\\
    \mathcal{L}_{o} &= \sum_{\mathbf{p}'} \ell_1(\dot{\mathbf{J}}_o(\mathbf{p}'), \dot{\mathbf{J}}_o^{gt}(\mathbf{p}'))\cdot \dot{\mathbf{J}}_{pt}^{gt}(\mathbf{p}'),
\end{align}

The total loss of learning endpoints is $\mathcal{L}_{\dot{\mathbf{J}}}=\beta_{pt}\cdot\mathcal{L}_{pt}+\beta_0\cdot\mathcal{L}_{o}$, where $\beta_{pt}$ and $\beta_o$ are trade-off hyperparameters. In our experiments,  $\beta_{pt}$ and $\beta_o$ are set to 8.0 and 0.25, {exactly same to L-CNN~\cite{L-CNN} and our preliminary version, HAWPv1~\cite{HAWP}.}

\paragraph*{Extracting Endpoints from the Dense Predictions.} With the predicted heatmap scores, we first apply the local NMS using a $3\times 3$ window, and then keep the top-$N$ endpoints using a sufficiently large number $N$. In our experiments, we use $N = \max(2\times N_{gt}, 300)$ in the training phase where $N_{gt}$ the number of junctions in an image. In testing, we use $N=\max(N_{pred}, 300)$ with $N_{pred} = \sum_{\mathbf{p}'} \mathbf{1}_{\dot{\mathbf{J}}_{pt}(\mathbf{p}')\geq \tau_{\jmath}}$, {where $\tau_j$ is set to $0.008$, same as L-CNN~\cite{L-CNN} and our HAWPv1~\cite{HAWP}}. 
For the top-$N$ endpoints, their (sub-pixel) locations are updated based on the predicted offsets in $\dot{\mathbf{J}}_{o}$. %

\subsection{Binding Line Segment and Endpoint Proposals}\label{sec:binding}

{Denote by $\hat{\ddot{\mathbf{L}}}$  the set of line segment proposals from the HAT field prediction and by $\hat{\dot{\mathbf{J}}}$ the set of endpoint proposals. Binding them captures their co-occurrence and improves the fidelity of line segment proposals while significantly reducing the computational cost to handle the total of $|\hat{\ddot{\mathbf{L}}}| = (2k+1)\cdot H_s\cdot W_s$ proposals in $\hat{\ddot{\mathbf{L}}}$, particularly those strongly supported by endpoint proposals.

Specifically, we first find the nearest endpoint proposals in $\hat{\dot{\mathbf{J}}}$ for the two endpoints of a line segment proposal in $\hat{\ddot{\mathbf{L}}}$. Without loss of generality,  consider a line segment proposal $\hat{\ddot{l}}=(\hat{\mathbf{x}}_1, \hat{\mathbf{x}}_2)$, denote by $\hat{\mathbf{y}}_1 \in \hat{\dot{\mathbf{J}}}$ the nearest endpoint proposal for $\hat{\mathbf{x}}_1$ and {the squared} Euclidean distance between them is $\delta_1$. The same is done for $\hat{\mathbf{x}}_2$ and we obtain $\hat{\mathbf{y}}_2$ and $\delta_2$. The figure of merit of the binding is defined by the maximum distance $\delta=\max(\delta_1, \delta_2)$, and the smaller the distance, the higher the quality of the line segment $\hat{\ddot{l}}$ (\ie, the likelihood of the co-occurrence is high). 
A threshold $\tau_{\delta}$ is used to select proposals of high-quality line segments whose binding costs $\delta$ are smaller than the threshold {($\tau_{\delta}={10}$} in our experiments). Based on this, the $|\hat{\ddot{\mathbf{L}}}|$ number of line segments $\hat{\ddot{\mathbf{L}}}$ in are anchored by the set $\hat{\dot{\mathbf{J}}}$ in a sparse set without incurring any non-maximal suppression schema of line segments, and eventually yield a new set of \textbf{endpoint-augmented line segment} proposals consisting of line segments $\hat{\ddot{l}}_j=(\hat{\mathbf{y}}_1, \hat{\mathbf{y}}_2; \hat{\mathbf{x}}_1, \hat{\mathbf{x}}_2)$, where $\hat{\ddot{l}}_j$ is used to indicate that the line segment is formed by the binding process, and the endpoints from both $\hat{\ddot{\mathbf{L}}}$ and $\hat{\dot{\mathbf{J}}}$ are retained for preserving the instrinsic uncertainty of proposals. Eventually, the endpoint-augmented line segment proposals generate a new set $\hat{\ddot{\mathbf{L}}}_{\dot{\mathbf{J}}}$. 
}

\subsection{Line Segment Verification}\label{sec:loi-pooling}
Recall that in the proposed HAT field, points on line segments are treated as ``background" points. Even after the binding, those points have not been verified. Thus, verifying line segment proposals in $\hat{\ddot{\mathbf{L}}}_{\dot{\mathbf{J}}}$ entails grounding the entirety of a line segment proposal to the data evidence, \ie, Line-of-Interest (LOI) Pooling, to compute the final score for wireframe parsing, where the entirety is defined with respect to a point-sampling strategy. We give details of our proposed endpoint-decoupled (EPD) LOIAlign method, which is built on the vanilla LOIPooling method~\cite{L-CNN}. 

To enable an efficient design of the verification head MLP classifier, we ask the question: \textit{Do we need to encode every sampled point in the same high-dim space as done in the vanilla LOIPooling?} By definition, the two endpoints of a line segment play a critical role. So, we propose to encode the two endpoints and the sampled intermediate points of a line segment differently in a high-dim feature space and a low-dim feature space, respectively, that is, to develop the EPD LOIAlign.   

Denoted by a linear-sampling function $\psi_t(\lseg)$ that maps a line segment $\ddot{l} = (\mathbf{x}_1, \mathbf{x}_2)$ to a point in the line segment by
\begin{equation}\label{eq:loi-sampling}
    \psi_t(\mathbf{x}_1,\mathbf{x}_2) = (1-t)\cdot\mathbf{x}_1 + t\cdot\mathbf{x}_2, \,t\in [0,1],
\end{equation}
where a predefined number of evenly-sampled points $t_i=\frac{i}{n}$ is typically used (\eg, $n=31$ and $i\in \{0,1,\cdots n\}$). 

Consider an endpoint-augmented line segment $\hat{\ddot{l}}_j=(\hat{\mathbf{y}}_1, \hat{\mathbf{y}}_2; \hat{\mathbf{x}}_1, \hat{\mathbf{x}}_2)$, for the LOI verification, we maintain three subsets of sampled points: 
\begin{itemize}
    \item The two endpoints: $\{(\hat{\mathbf{y}}_1, \hat{\mathbf{y}}_2)\}$; %
    \item The intermediate points between $\hat{\mathbf{y}}_1$ and $\hat{\mathbf{y}}_2$: $\mathcal{Y} = \{\psi_{t_i}^y = \psi_{t_i}{(\hat{\mathbf{y}}_1, \hat{\mathbf{y}}_2)}); i=1,2,\cdots,n-1\}$;%
    \item The intermediate points between $\hat{\mathbf{x}}_1$ and $\hat{\mathbf{x}}_2$: $\mathcal{X} = \{\psi_{t_i}^x = \psi_{t_i}{(\hat{\mathbf{x}}_1, \hat{\mathbf{x}}_2)}), i=1,2,\cdots,n-1\}$.%
\end{itemize}
{
By decoupling endpoints and intermediate points, the model will be geometrically aware for line segments in learning to verify the proposals. We note that our proposed EPD LOIAlign captures the co-occurrence between the HAT-drive inductive line segment proposals and the junction-guided deductive line segment proposals for better verification. 

\paragraph*{The Verification Head Classifier.} 
To exploit the EPD LOIAlign and to enrich the information flow for the low-dimensional feature extractor $f_{\psi}$, we utilize a parallel branch design of the verification head classifier. As shown in Fig.~\ref{subfig:verification}, we use three convolution layers to transform the backbone network to $F_J \in \mathbb{R}^{C\times H_s\times W_s}$ for the two endpoints, as well as $F_{\mathcal{Y}} \in \mathbb{R}^{C_{\psi}\times H_s\times W_s}$ and $F_{\mathcal{X}} \in \mathbb{R}^{C_{\psi}\times H_s\times W_s}$ for the sampled point set $\mathcal{Y}$ and $\mathcal{X}$. Note that the feature maps $F_{\mathcal{Y}}$ and $F_{\mathcal{X}}$ have fewer channels than $F_J$ to deal with the sample points $(n-1)$ for each proposal. The bilinear interpolation is used to sample all features, and yield three feature vectors for each proposal, $\mathbf{Z}_{\mathbf{y}}(\hat{\lseg}_j)$ for the two endpoints, as well as $\mathbf{Z}_{\psi_{\mathbf{y}}}$ and $\mathbf{Z}_{\psi_{\mathbf{x}}}$  for the sampled point set $\mathcal{Y}$ and $\mathcal{X}$ for the line verification, denoted by 
\begin{align}
    \mathbf{Z}_{\mathbf{y}}(\hat{\lseg}_j) = [F_J(\mathbf{y}_1),F_J(\mathbf{y}_2)] \in \mathbb{R}^{2C},\\
    \mathbf{Z}_{\psi_{\mathbf{y}}}(\hat{\lseg}_j) = [F_{\mathcal{Y}}(\psi_{t_1}^y ),\ldots,F_{\mathcal{Y}}(\psi_{t_{n-1}}^y)] \in \mathbb{R}^{(n-1)\cdot C_{\psi}},\\
    \mathbf{Z}_{\psi_{\mathbf{x}}}(\hat{\lseg}_j) = [F_{\mathcal{X}}(\psi_{t_1}^x ),\ldots,F_{\mathcal{X}}(\psi_{t_{n-1}}^x)] \in \mathbb{R}^{(n-1)\cdot C_{\psi}}.
\end{align}
Let $\mathbf{Z}_{\psi}(\hat{\ddot{l}}_j)=[\mathbf{Z}_{\psi_\mathbf{y}}(\hat{\ddot{l}}_j), \mathbf{Z}_{\psi_\mathbf{x}}(\hat{\ddot{l}}_j)]\in \mathbb{R}^{2\cdot(n-1)\cdot C_{\psi}}$ be the concatenated features of the intermediated sample points and $\mathbf{Z}(\hat{\ddot{l}}_j)=[\mathbf{Z}_{\mathbf{y}}(\hat{\ddot{l}}_j), \mathbf{Z}_{\psi}(\hat{\ddot{l}}_j)]\in \mathbb{R}^{2\cdot(n-1)\cdot C_{\psi}+2\cdot C}$ be the concatenated features of all sample points, we first apply two separate MLPs to transform $\mathbf{Z}_{\psi}$ and $\mathbf{Z}$ into the $D$-dim feature space (\eg, $D=128$) respectively, and then sum them, followed by appling a linear transformation to compute the final score (or logit) of the line segment. For simplicity of notation, we will omit ``$(\hat{\ddot{l}}_j)$" and instead use $\mathbf{Z}_{\psi}$ and $\mathbf{Z}$.
\begin{equation}
    \text{Score}(\hat{\ddot{l}}_j) = \text{Linear}(\text{MLP}(\mathbf{Z}_{\psi}) + \text{MLP}(\mathbf{Z})). \label{eq:ver}
\end{equation}
To further enhance the learning of the low-dimensional feature extractor $f_{\psi}$, we use an auxiliary verification layer (linear transformation)  that is used in training only, 
\begin{equation}
    \text{AuxScore}(\hat{\ddot{l}}_j) = \text{Linear}(\mathbf{Z}_{\psi}). \label{eq:ver_aux}
\end{equation}
}

\paragraph*{The Ground-Truth Assignment in Training.} A line segment proposal  $\hat{\ddot{l}}_j=(\hat{\mathbf{y}}_1, \hat{\mathbf{y}}_2; \hat{\mathbf{x}}_1, \hat{\mathbf{x}}_2)$ is assigned as positive if and only if there exists a ground-truth line segment $\ddot{l}$ within the close proximity of $\hat{\ddot{l}}_j$. Similar to the binding between line segment proposals and endpoint proposals, the proximity is defined by the maximum distance between the two endpoints of the ground-truth line segment and $(\hat{\mathbf{y}}_1, \hat{\mathbf{y}}_2)$. By close, it means that the maximum distance needs to be smaller than a predefined threshold, $\tau_{ver}$ (\eg, $\tau_{ver}=1.5$ used in our experiments). This ground-truth assignment method facilitates learning the negatives on the fly and end-to-end, unlike previous work~\cite{HAWP,L-CNN} that exploit static negative line segment samples (\eg, by sampling two s that do not come from any ground-truth line segments). As we shall show, this on-the-fly and end-to-end generation of negative samples enables much more effective self-supervised learning of our HAWP model.

\paragraph*{Loss Functions.} The verification is posed as the binary classification problem. We utilize the BCE loss in training both the verification head classifier and the auxiliary head classifier. Denote by $\mathcal{L}_{ver}$ and $\mathcal{L}_{aux}$ the BCE loss functions for training Eq.~\eqref{eq:ver} and Eq.~\eqref{eq:ver_aux} respectively.

\section{Learning HAWP Models}\label{sec:models}
In this section, we present details of training and inference of both HAWPv2 and HAWPv3. Due to the space limit, we present network architectures used in our experiments in the Appx.~\ref{appx:networks}.

\subsection{HAWPv2 with Fully-Supervised Learning}\label{sec:fsl-setting}

\paragraph*{Training Details} 
Our proposed HAWPv2 is trained on the Wireframe dataset ~\cite{wireframe-dataset} with their official annotations. In the training, a total of $5000$ training samples with precisely annotated wireframes are augmented by the flipping operations along the horizontal, vertical, and diagonal directions, as well as the rotation operations\footnote{After 2020, the more complicated data augmentation techniques including the rotational augmentation used in F-Clip~\cite{FClip} and ELSD~\cite{ELSD}, the random cropping and color jittering used in LETR~\cite{LETR} were explored to boost the final performance.
} by $90^\circ$ and $-90^\circ$. Finally, the augmented dataset with 30k samples is used to train HAWPv2 with a total of 30 epochs. The ADAM optimizer~\cite{kingma2014adam} is used for training, and the learning rate is initially set at 4e-4 for the first 25 epochs and then reduced to 4e-5 for the last 5 epochs. The image resolution of the training samples is set to $512\times 512$. The total loss is the sum of the loss functions defined in Sec.~\ref{sec:framework}.

\paragraph*{Inference Details} Given an input image, we forward the trained HAWPv2 model to get the junctions and line segments. We only keep the line segments of which the classification score $c_i$ is greater than a given threshold $\varepsilon$ as the final prediction. The input image is also resized to $512\times 512$ tensors for the forward to obtain the wireframes that are then scaled to their original resolution according to the scaling factors in both horizontal and vertical directions.
\subsection{HAWPv3 with Self-Supervised Learning}

\begin{figure}[!t]
    \centering
    \subfigure[Checkboard]{
    \includegraphics[width=0.2\linewidth]{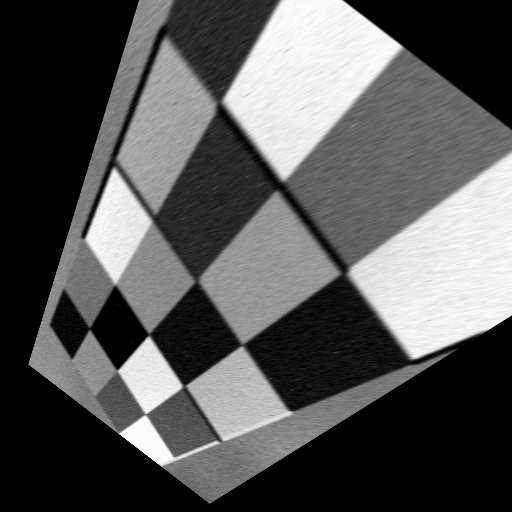}
    }
    \subfigure[Lines]{
    \includegraphics[width=0.2\linewidth]{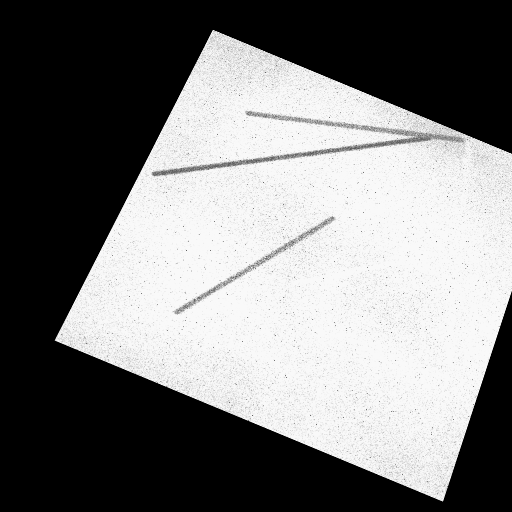}
    }
    \subfigure[A Cube]{
    \includegraphics[width=0.2\linewidth]{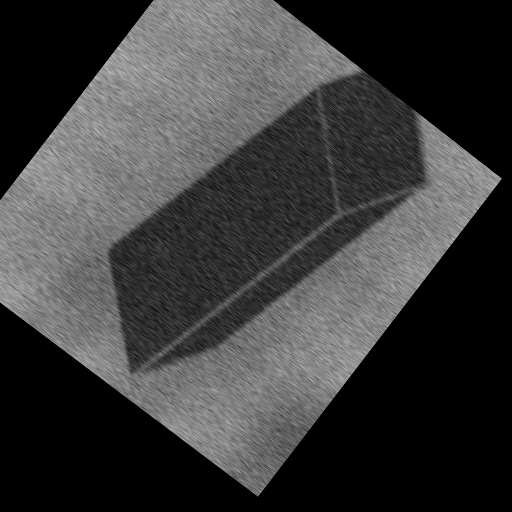}
    }
    \subfigure[Gaussian]{
    \includegraphics[width=0.2\linewidth]{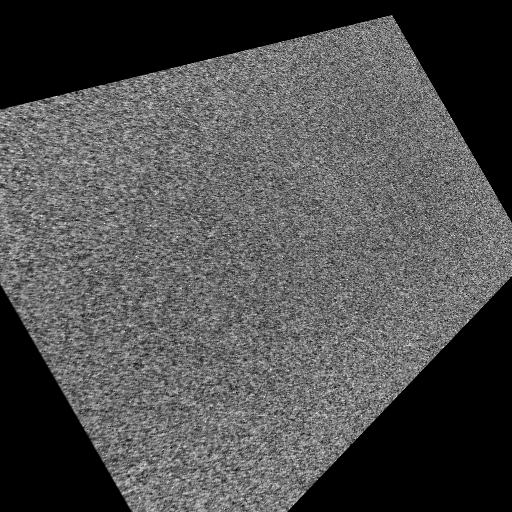}
    }
    
    \subfigure[Strips]{
    \includegraphics[width=0.2\linewidth]{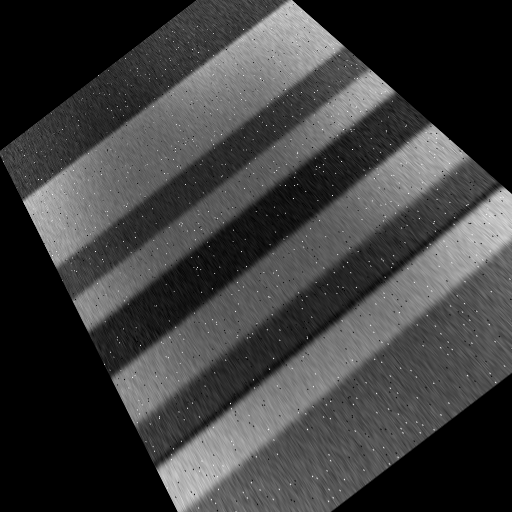}
    }
    \subfigure[A Polygon]{
    \includegraphics[width=0.2\linewidth]{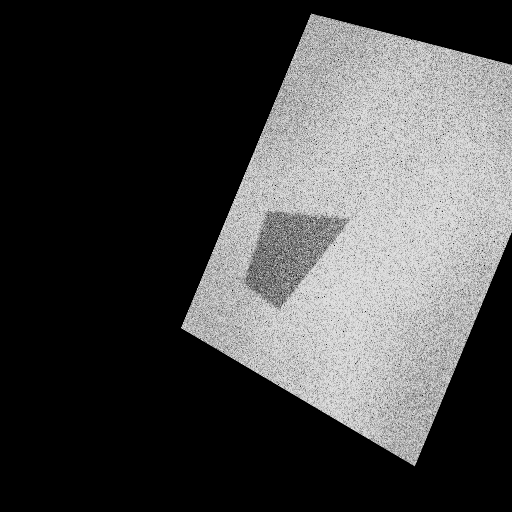}
    }
    \subfigure[Polygons]{
    \includegraphics[width=0.2\linewidth]{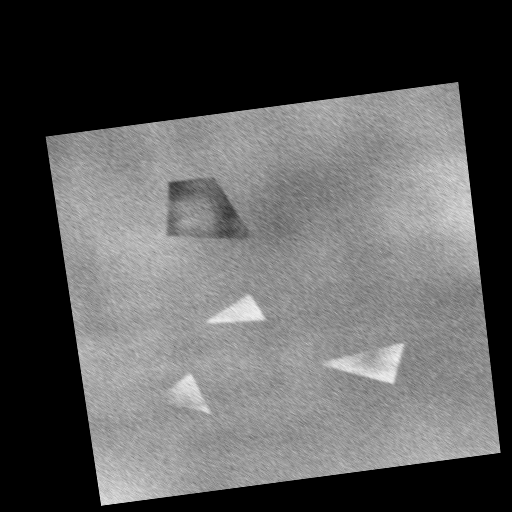}
    }
    \subfigure[A Star]{
    \includegraphics[width=0.2\linewidth]{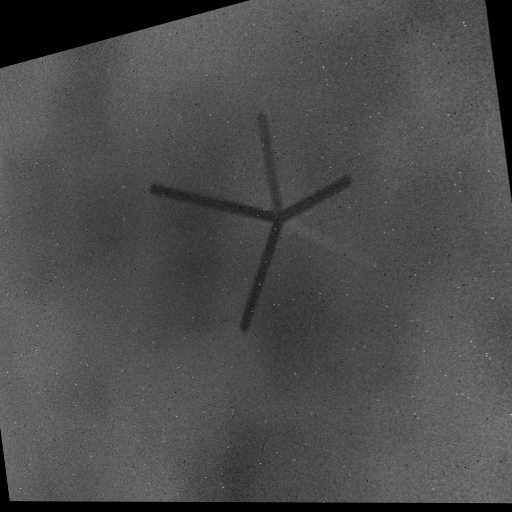}
    }
    \caption{Some training examples generated in the synthetic dataset for the initialization of self-supervised learning. There are 8 primitives in total used in the dataset and we show  one random synthesized image for each primitive.}
    \label{fig:synthetic-data}
\end{figure}

\begin{figure}[!t]
    \centering
    \includegraphics[width=\linewidth]{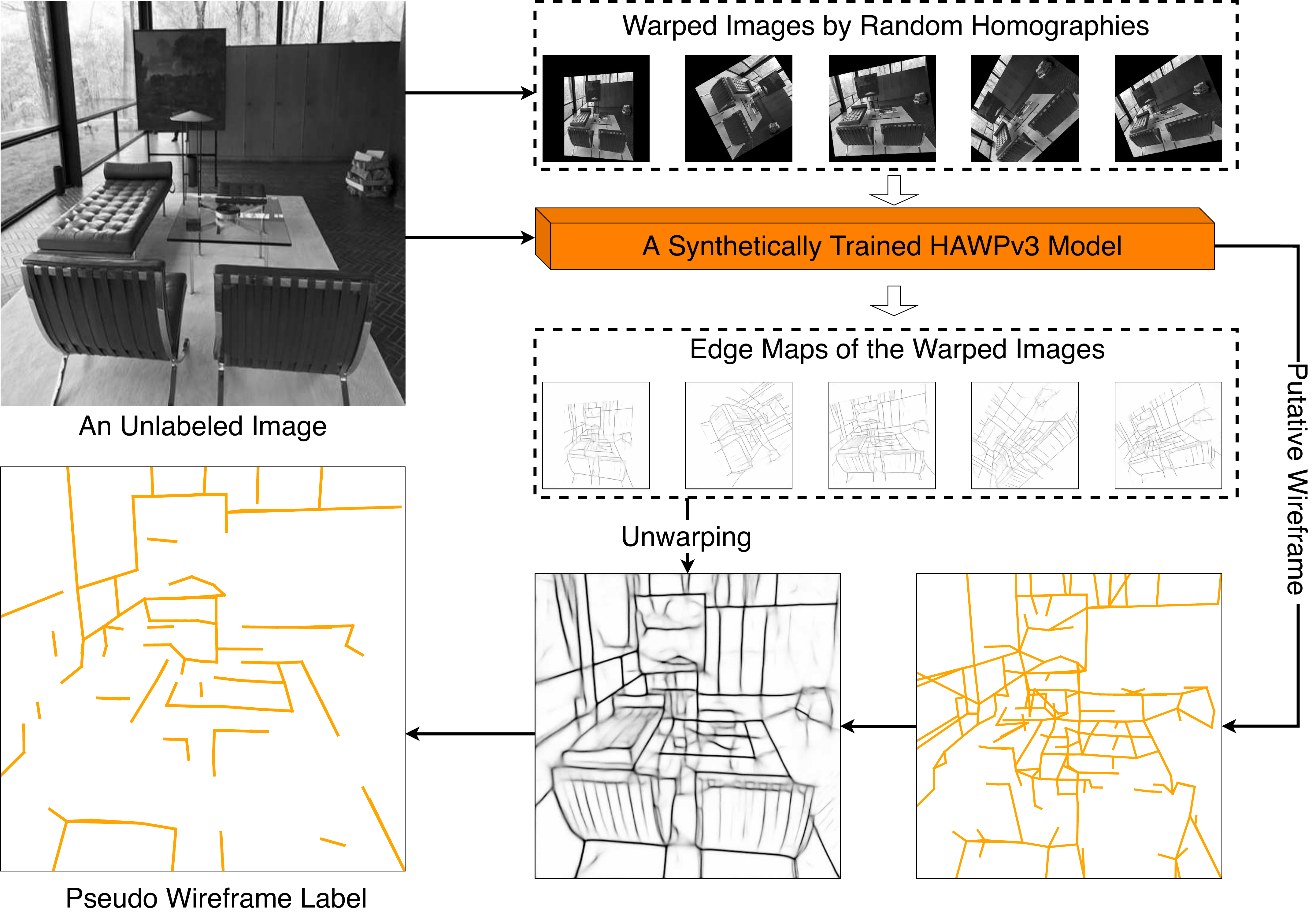}
    \caption{
        The homographic adaptation scheme for the HAWPv3 training. Different from SOLD$^2$~\cite{SOLD2}, in our HAWPv3, the warped images are only used to predict the edge maps that are then unwarped back to the original view and yield an averaged edge prediction. The edge predictions are treated as additional evidence to prune out the outliers of line segments in the putative wireframe computed from the original unlabeled image.
    }
    \label{fig:had-scheme}
\end{figure}

We adopt the simulation-to-reality pipeline for SSL as in the SuperPoint~\cite{SuperPoint} and the SOLD$^2$~\cite{SOLD2}. In the synthetic pretraining using simulated data consisting of simple primitives (Fig.~\ref{fig:synthetic-data}), ground-truth wireframes are naturally available. We can thus train our HAWPv2 model. The challenge is how to leverage the synthetically pretrained HAWPv2 to ``annotate" wireframes in real images. If we directly run the synthetically trained HAWPv2 on an input unlabeled image. The putative wireframe is often quite noisy due to the gap between synthetic data and real images. To address this challenge, we need to introduce additional inductive biases of line segments that are more transferrable from simulation to reality, such that we can exploit them to ``clean up" the putative wireframe to get the pseudo wireframe labels for the input image. The resulted pseudo wireframe labels will be used in training a HAWP model from scratch.  We utilize the Homography Adaptation method~\cite{SuperPoint} (Fig.~\ref{fig:had-scheme}) and resort to learn the edge map as cues in the ``cleaning-up" process. This leads to our design of HAWPv3.  

{
Our HAWPv3 is built on the HAWPv2 with a new sub-network added for high resolution predictions of edge and junction heatmaps as well as the junction offsets, denoted by $f_{\rm edge}(F;\Omega_{\rm edge})$, $f_{\rm jhm}(F;\Omega_{\rm jhm})$ and $f_{\rm joff}(F;\Omega_{\rm joff})$,
\begin{align}
    f_{\rm edge}: \quad \text{ReLU}\circ\text{Conv}_{1\times1}^{\frac{C}{s^2}\to 1}\circ\text{PixelShuffle}, \label{eq:edge-pred}  \\
    f_{\rm jhm}: \quad \text{ReLU}\circ\text{Conv}_{1\times1}^{\frac{C}{s^2}\to 1}\circ\text{PixelShuffle}, \label{eq:jucs-pred}  \\
    f_{\rm joff}: \quad \text{ReLU}\circ\text{Conv}_{1\times1}^{\frac{C}{s^2}\to 2}\circ\text{PixelShuffle}, \label{eq:jucs-off-pred}  
\end{align}
where the PixelShuffle~\cite{shi2016real} operation rearranges elements in the feature map $F$ of shape $(C, H_s, W_s)$ to a feature map of shape $(\frac{C}{s^2}, H, W)$. Recall that $s$ is the overall stride used in computing $F$, the predictions are at the same resolution as the input image (\eg, $512\times 512$). With the high resolution prediction of junctions, our HAWPv3 could handle localization errors better in practice.
}

The proposed HAWPv3 is first trained using synthetic data. For the edge branch, we use the balanced cross-entropy loss function in training, together with other loss functions defined in training HAWPv2.

\paragraph*{Synthetic Pretraining of HAWPv3.} We generate 2,000 images for each of the eight primitives (Fig.~\ref{fig:synthetic-data}) and obtain 16,000 images  in total. We train our HAWPv3 with 10 epochs.  The learning rate is set to 4e-4 constantly over the entire training schedule. Compared to SOLD$^2$~\cite{SOLD2} that used $10$ times more synthetic images (\ie, 160k images) and 200 epochs of training, HAWPv3 is much more efficient.

\paragraph*{``Annotating" Real Images using HAWPv3 with Homography Adaptation.} 
In Fig.~\ref{fig:had-scheme}, we use the synthetically trained HAWPv3 to annotate real images. Given an unlabeled real image $I$, we randomly warp it using $N_h$ homographies ($N_h=10$). The warped images are used to compute edge maps, which are then unwarped and averaged to generate the predicted edge map for $I$. From the original image $I$, we compute the putative wireframe. During verification, we combine the EPD LOIAlign score ($c_i$) and the edge map score ($c_i'$) for each line segment ($i$). The edge map score is computed by sampling and averaging local maximal edge scores for 64 points on the line segment.
The final SSL score for a line segment is the harmonic average, $c^{SSL}_i=\sqrt{c_i\cdot c'i}$. We prune line segments with scores below the threshold $\tau_{SSL} = 0.75$ in the putative wireframe. Unlike SOLD$^2$~\cite{SOLD2}, which uses adapted edge maps and end-point heatmaps, we only use the former. We employ 20k training images for SSL, generated from the original wireframe dataset~\cite{wireframe-dataset} by flipping horizontally, vertically, and diagonally.

\paragraph*{Training HAWPv3 with Pseudo Wireframe Labels.} With the pseudo wireframe labels, training HAWPv3 is supervised learning again, as done in training our HAWPv2. We train a new HAWPv3 model from scratch with 30 epochs following the same settings as HAWPv2. We do not use the model weights from the synthetic pretraining due to the semantic gap between the synthetic data and real images.  We notice that the process of ``annotating" real images can be applied in a cascade manner, that is to use  the pseudo-wireframe trained HAWPv3 to re-compute  putative wireframes and edge maps to update/refine pseudo wireframes for a new round of SSL (see Sec.~\ref{sec:ssl-exp} and the  Appx.~\ref{appx:detailed-hawpv3} for the experimental results and detailed analyses).

\section{Experiments of HAWPv2}\label{sec:exp-fsl}

In this section, we test the proposed HAWPv2 and compare it with state-of-the-art line segment detectors and wireframe parsers in the fully supervised learning (FSL) setting. All methods are trained in the Wireframe training dataset~\cite{wireframe-dataset}, and tested with the testing samples (462 images in total) of the Wireframe dataset~\cite{wireframe-dataset} and the entire YorkUrban dataset (102 images in total)~\cite{Denis2008}. 
We also present ablation studies to verify the design of the proposed HAWPv2.

\begin{table*}
    \caption{Quantitative results and comparisons. Our proposed HAWPv2 sets new records in the most challenging metrics of structural correctness. For heatmap-based evaluation results, our proposed HAWPv2 obtains better average precision and performance comparable to heatmap-based F scores. In the last column, the inference speed is compared for all learning-based approaches. The numbers with $^*$ are extracted from the original paper.  The best scores are highlighted in {\bf bold fonts}.
    }
    \centering
    \resizebox{0.95\linewidth}{!}{ 
    \begin{tabular}{r|c|c|ccc|c|l|l|ccc|c|l|l|l}
        \toprule
        & \multirow{2}{*}{Image Size} & \multirow{2}{*}{\# Epochs} & \multicolumn{6}{c|}{\textit{Wireframe Dataset}} & \multicolumn{6}{c|}{\textit{YorkUrban Dataset}}& \multirow{2}{*}{FPS}
        \\\cline{4-15}
        &            & & sAP$^5$ & sAP$^{10}$ & sAP$^{15}$  & mAP$^{J}$ & AP$^{{H}}$ & F$^{{H}}$ & sAP$^5$ & sAP$^{10}$ & sAP$^{15}$  & mAP$^J$ & AP$^{{H}}$ & F$^{{H}}$  &
        \\\midrule
        DWP~\cite{wireframe-dataset} & 320$\times$320 & 200 & 3.7 & 5.1 & 5.9 & 40.9 & 67.8 & 72.2  & 1.5 & 2.1 & 2.6 & 13.4 & 51.0 & 61.6 & 2.24\\\midrule
        AFM~\cite{AFM-CVPR} & 320$\times$320 & 200 & 18.5 & 24.4 & 27.5 & 23.3 & 69.2 & 77.2 & 7.3 & 9.4 & 11.1 & 12.4 &48.2 & 63.3 & 13.5 \\\midrule
        AFM++~\cite{RegionalAttraction} & 512$\times$512 & 200 & 27.7 & 32.4 & 34.8 & 30.8 & 74.8 & {\bf 82.8} & 9.5 & 11.6 & 13.2 & TBD & 50.5 & {\bf 66.8} & 5.2\\\midrule
        L-CNN~\cite{L-CNN} & 512$\times$ 512 & 30 & 59.7 & 63.6 & 65.3 & 60.2 & 81.6 & 77.9 & 25.0 & 27.1 & 28.3 & 31.5 & 58.3 & 62.2 & 15.6 \\\midrule
        F-Clip (HG2-LB)~\cite{FClip} & 512$\times$ 512 & 300 & 62.6 & 66.8 & 68.7 & 48.7 & 85.1 & 80.9 &  27.6 & 29.9 & 31.3 & 28.3 & 62.3 & 64.5 & 28.3\\\midrule
        LETR (R101)~\cite{LETR} & 800 (short) & 825 & 59.2 & 65.2 & 67.7  &  44.1  &  85.5    & 79.8  & 23.9 & 27.6 & 29.7 & 24.5 & 59.6 & 62.0 & 5.25 \\
        LETR (R50) ~\cite{LETR} & 800 (short) & 825 & 58.5 & 64.6 & 67.3  &  44.2 &  84.7    & 79.1  & 25.7 & 29.6 & 32.0 &  25.9 & 61.7 & 63.4 & 5.25 \\\midrule
        ELSD (HG)~\cite{ELSD} & \multirow{2}{*}{512$\times$ 512} & 170 & 62.7$^*$ & 67.2$^*$ & 69.0$^*$ & N/A & 84.7$^*$ & 80.3$^*$ & 23.9$^*$ & 26.3$^*$ & 27.9$^*$ & N/A & 57.8$^*$ & 62.1$^*$ & 47$^*$ \\
        ELSD (Res34)~\cite{ELSD} &                               & 170 & 64.3$^*$ &  68.9$^*$    & 70.9$^*$ & N/A & 87.2$^*$ & 82.3$^*$ & 27.6$^*$ & 30.2$^*$ & 31.8$^*$ & N/A & 62.0$^*$ & 63.6$^*$ & 42.6$^*$
        \\\midrule
        HAWPv1 (Ours)~\cite{HAWP} & \multirow{2}{*}{512$\times$ 512} & 30 & 62.5 & 66.5 & 68.2 & 60.2 & 84.5 & 80.3 & 26.1 & 28.5 & 29.7 & 31.6 & 60.6 & {64.8} & 29.5\\
        HAWPv2 (Ours) & & 30 & \textbf{65.7}  & \textbf{69.7} & \textbf{71.3} & {\bf 61.8} & {\bf 88.0} & 81.4 & \textbf{28.8} & \textbf{31.2} & \textbf{32.6} & {\bf 32.5} & {\bf 64.6} & 64.5 & 40.8 \\
        \bottomrule
    \end{tabular}
    }
    \vspace{-3mm}
    \label{tab:summary-metric}
\end{table*}
\subsection{Evaluation Metrics}
\paragraph*{{Structural Average Precision (sAP)}.} This is motivated by the typical AP metric used in evaluating object detection systems, and has been the most challenging metric for wireframe parsing and line segment detection (LSD). A counterpart of the Intersection-over-Union (IoU) overlap is used. For each ground-truth line segment $\ddot{l}=(\mathbf{x}_1, \mathbf{x}_2)$, we first find the set of parsed line segments each of which, $\hat{\ddot{l}}=(\hat{\mathbf{x}}_1, \hat{\mathbf{x}}_2)$, satisfies the ``overlap", 
\begin{equation}\label{eq:sap-dis}
    \min_{(i,j)} \left\|\mathbf{x}_1 - \hat{\mathbf{x}}_i \right\|^2 + \left\|\mathbf{x}_2 - \hat{\mathbf{x}}_j \right\|^2 \leq \vartheta_{L},
\end{equation}
where $(i,j)$ can be either $(1,2)$ or $(2,1)$, and $\vartheta_{L}$ is a predefined threshold. If no overlap is found, it is a False Negative (FN). If multiple candidates exist, the one with the highest verification score is a True Positive (TP), and the rest are False Positives (FP). Unmatched parsed line segments are also counted as FPs. We follow the convention used in previous methods to resize predictions and grountruth wireframes to $128\times 128$, and report sAP scores with thresholds $\vartheta$ of $5, 10, 15$, \ie, sAP$^{5}$, sAP$^{10}$, and sAP$^{15}$, respectively.

\paragraph*{{Heatmap based F score and Average Precision}.} These are traditional metrics used in LSD and wireframe parsing~\cite{wireframe-dataset}. Instead of directly using the vectorized representation of line segments, heatmaps are used, which are generated by rasterizing line segments for both parsing results and the ground truth. The pixel-level evaluation is used to calculate the precision and recall curves with which the heatmap F score, indicated by F$^H$ and the heatmap average precision, indicated by AP$^H$ are computed. Unlike the evaluation protocol that computes the F scores and AP by averaging the per-image evaluation result in~\cite{wireframe-dataset, AFM-CVPR, RegionalAttraction}, we follow L-CNN~\cite{L-CNN} to first calculate the true positive and false negative edge pixels over the entire dataset and then compute the F scores and AP.

\paragraph*{{Vectorized Junction Mean AP}.} It is calculated in a similar way to the sAP of line segments. Let $\vartheta_J$ be the threshold for the distance between a predicted junction and a ground truth one. The mAP$^J$ is computed~\emph{w.r.t.}~$\vartheta_J=0.5, 1.0, 2.0$. For the line segment detection approaches compared that did not yield junctions, we take the endpoints as the detected junctions for evaluation as in L-CNN~\cite{L-CNN}. 

\begin{figure*}[!ht]
    \centering
    \includegraphics[width=0.22\linewidth]{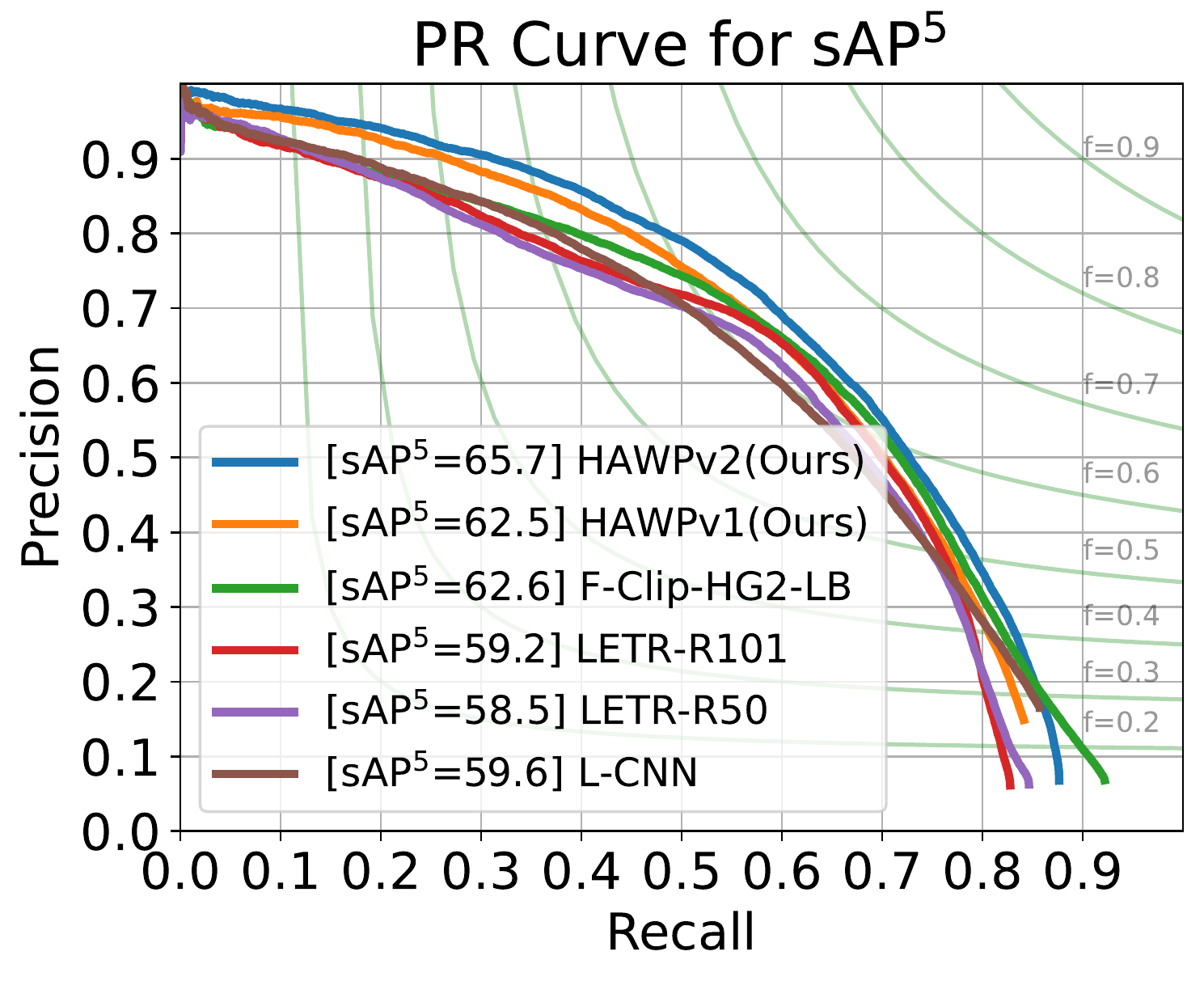}
    \includegraphics[width=0.22\linewidth]{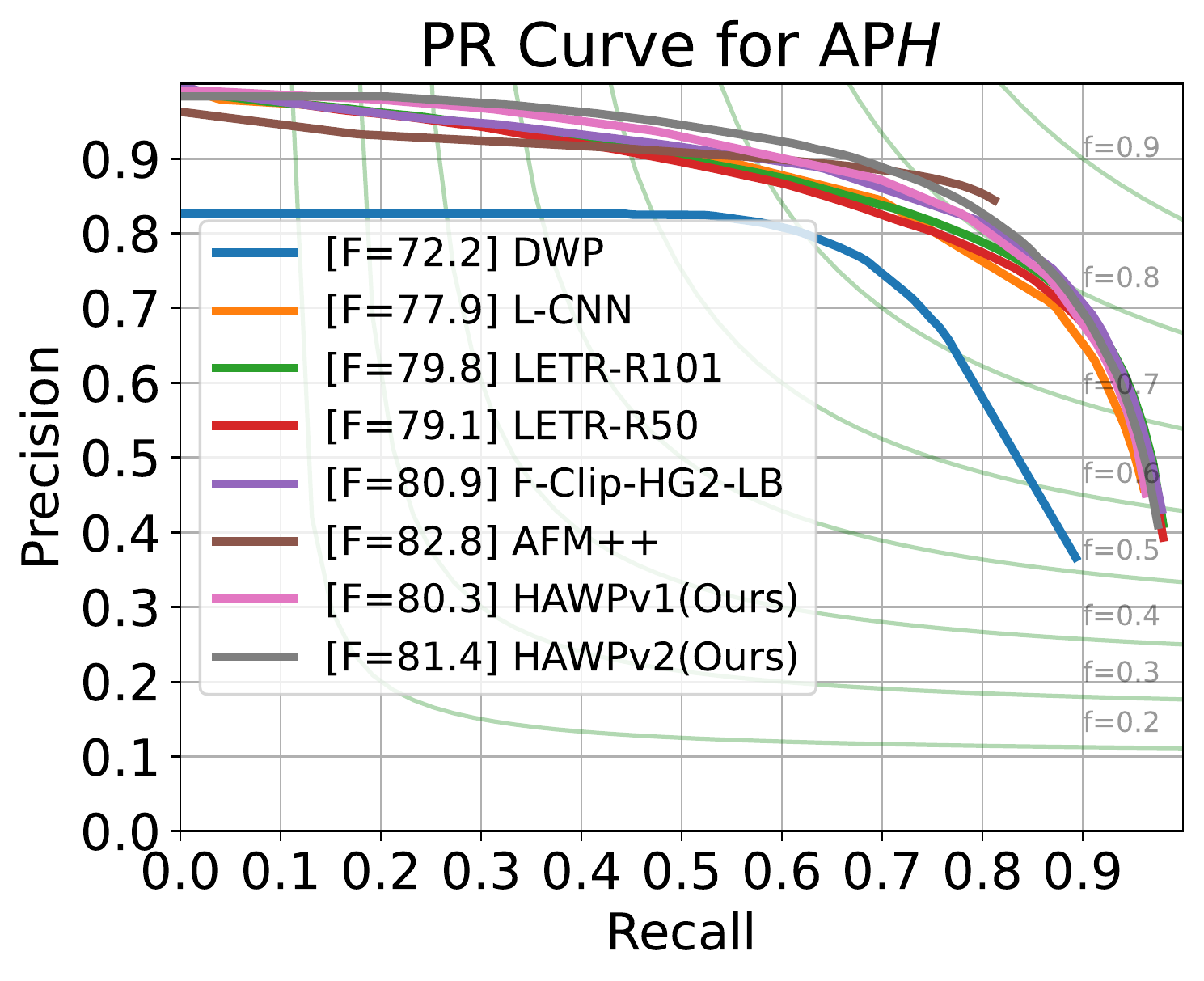}
    \includegraphics[width=0.22\linewidth]{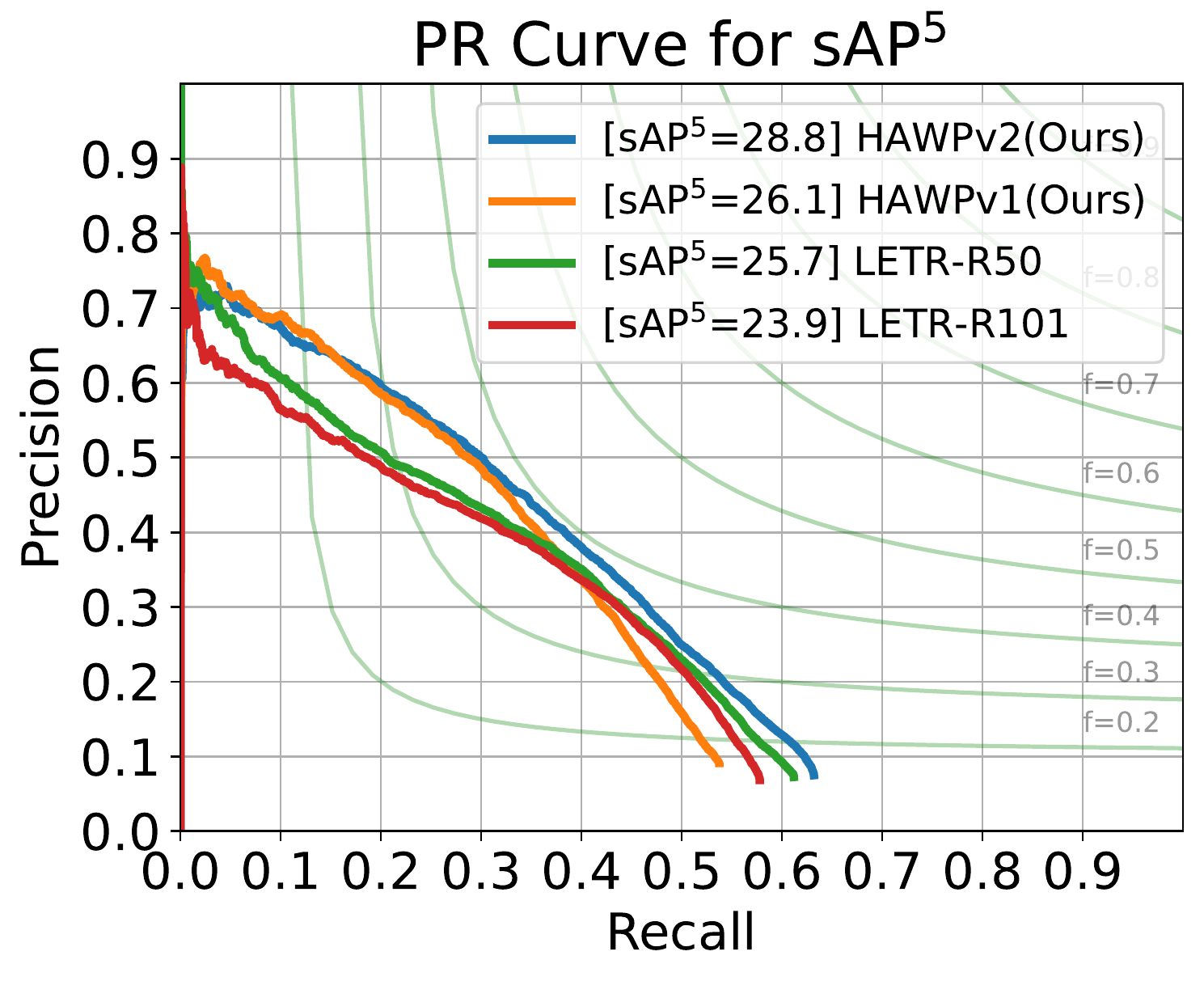}
    \includegraphics[width=0.22\linewidth]{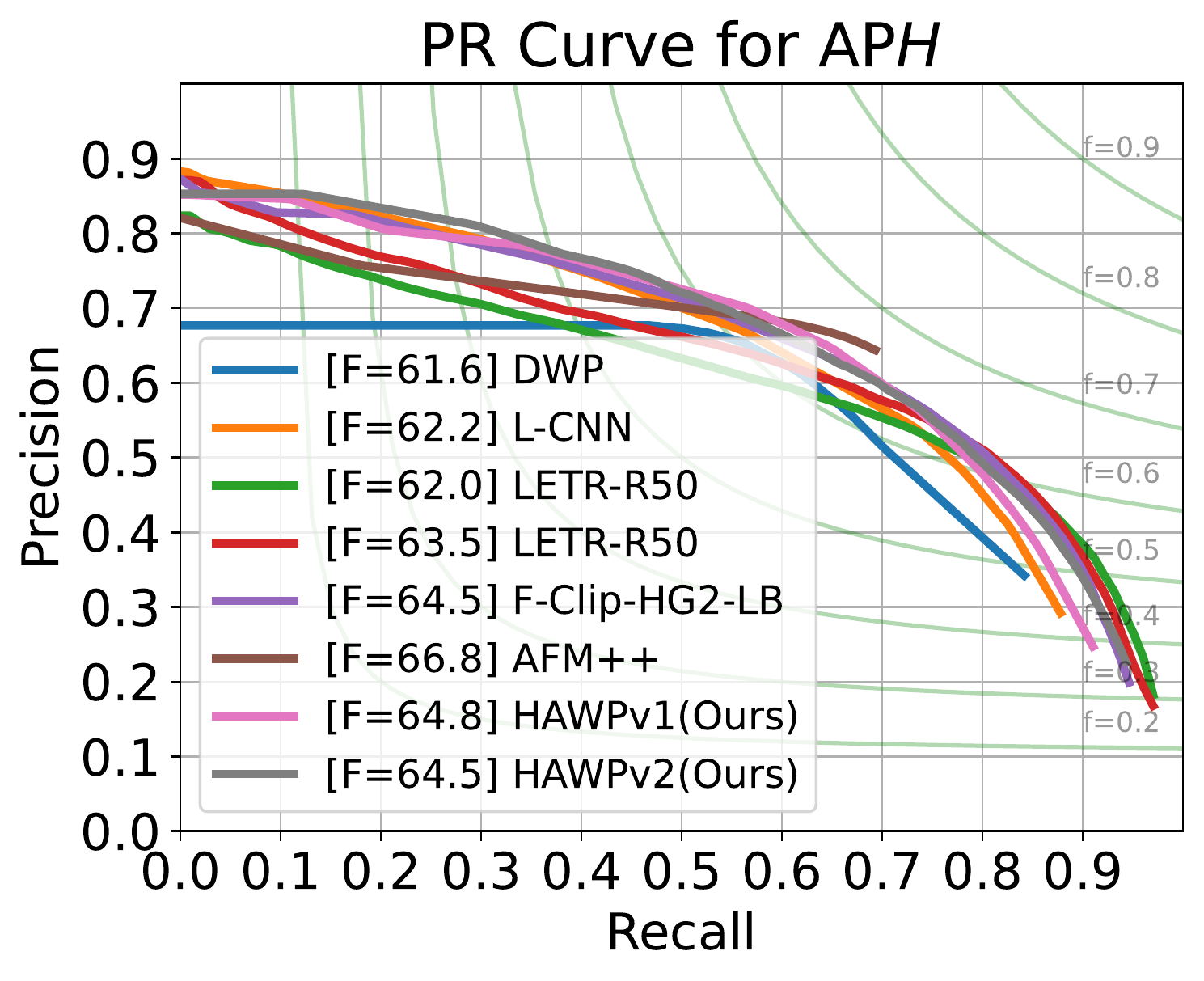}
    \vspace{-3mm}
    \caption{Precision-Recall Curves for sAP$^{5}$ and AP$^H$ on the Wireframe dataset (in the top row) and the YorkUrban dataset (in the bottom row)}
    \vspace{-5mm}
    \label{fig:pr-curves}
\end{figure*}
\subsection{Main Comparisons with State of the Arts}

\subsubsection{Baselines} 
We compare with the recent deep learning based approaches which are summarized as follows: 
\begin{enumerate}

    \item \emph{In 2018}, the Wireframe dataset~\cite{wireframe-dataset} was proposed with a baseline wireframe parser, DWP, which groups the learned junctions and line heatmaps into vectorized wireframe graphs. We use the DWP as the earliest baseline for wireframe parsing.
    
    \item \emph{In 2019}, there are several line segment detectors including AFM~\cite{AFM-CVPR} and AFM++~\cite{RegionalAttraction}, PPG-Net~\cite{PPGNet} and L-CNN~\cite{L-CNN}. The AFM approaches~\cite{AFM-CVPR,RegionalAttraction} are not fully end-to-end, while PPG-Net~\cite{PPGNet} and L-CNN~\cite{L-CNN} use neural networks to obtain the final wireframe graphs end-to-end. We compare our HAWPv2 with the AFM approaches~\cite{AFM-CVPR,RegionalAttraction} and L-CNN~\cite{L-CNN}.
    
    \item \emph{After 2019}, many works focused on the direct regression of line segments by learning center-based representations or exploiting Attention mechanisms in line segment detection. All these approaches require long learning schedules with hundreds of training epochs. We choose the best performing approaches, ELSD~\cite{ELSD}, F-Clip~\cite{FClip} and LETR~\cite{LETR} for comparison.
\end{enumerate}

\begin{figure*}[!ht]
    \centering
    \begin{minipage}[t]{0.18\linewidth}
        \centering
        \includegraphics[width=\linewidth]{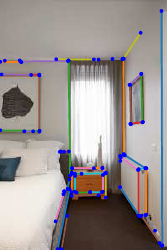}\\\vspace{2pt}
        \includegraphics[width=\linewidth]{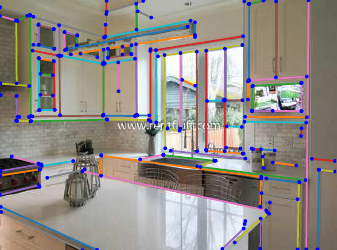}\\\vspace{2pt}
        \includegraphics[width=\linewidth]{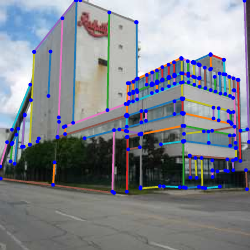}\\\vspace{2pt}
        \includegraphics[width=\linewidth]{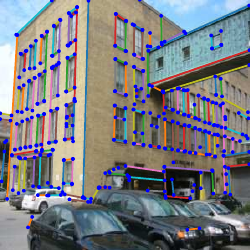}\\\vspace{2pt}
        AFM++~\cite{RegionalAttraction}
    \end{minipage}
    \begin{minipage}[t]{0.18\linewidth}
        \centering
        \includegraphics[width=\linewidth]{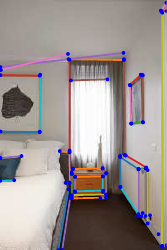}\\\vspace{2pt}
        \includegraphics[width=\linewidth]{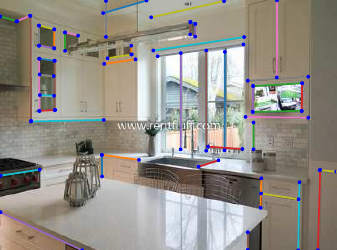}\\\vspace{2pt}
        \includegraphics[width=\linewidth]{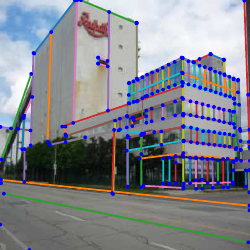}\\\vspace{2pt}
        \includegraphics[width=\linewidth]{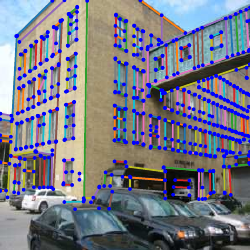}\\\vspace{2pt}
        F-Clip~\cite{FClip}
    \end{minipage}
    \begin{minipage}[t]{0.18\linewidth}
        \centering
        \includegraphics[width=\linewidth]{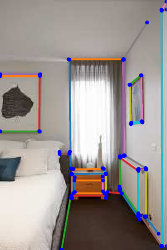}\\\vspace{2pt}
        \includegraphics[width=\linewidth]{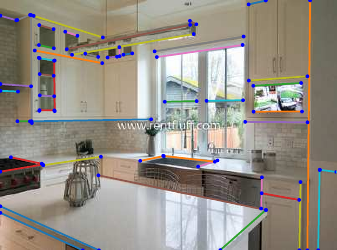}\\\vspace{2pt}
        \includegraphics[width=\linewidth]{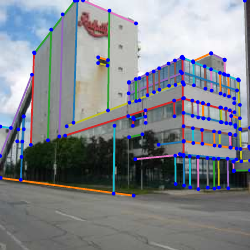}\\\vspace{2pt}
        \includegraphics[width=\linewidth]{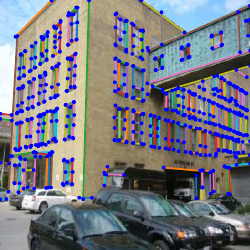}\\\vspace{2pt}
        LETR~\cite{LETR}
    \end{minipage}
    \begin{minipage}[t]{0.18\linewidth}
        \centering
        \includegraphics[width=\linewidth]{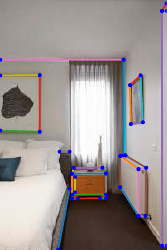}\\\vspace{2pt}
        \includegraphics[width=\linewidth]{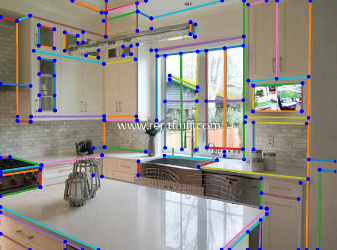}\\\vspace{2pt}
        \includegraphics[width=\linewidth]{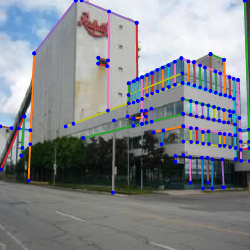}\\\vspace{2pt}
        \includegraphics[width=\linewidth]{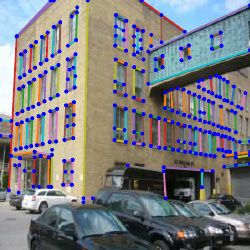}\\\vspace{2pt}
        HAWPv1 (Ours)~\cite{HAWP}
    \end{minipage}
    \begin{minipage}[t]{0.18\linewidth}
        \centering
        \includegraphics[width=\linewidth]{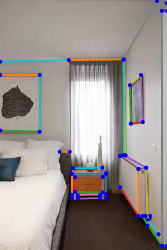}\\\vspace{2pt}
        \includegraphics[width=\linewidth]{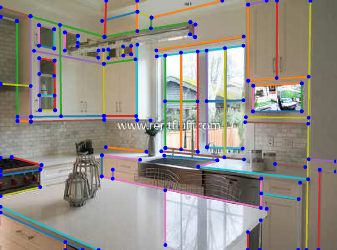}\\\vspace{2pt}
        \includegraphics[width=\linewidth]{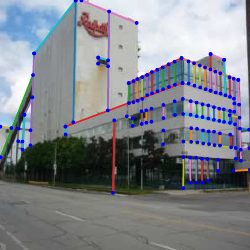}\\\vspace{2pt}
        \includegraphics[width=\linewidth]{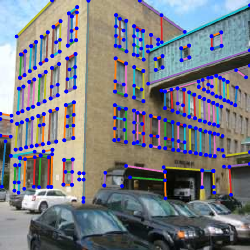}\\\vspace{2pt}
        HAWPv2 (Ours)
    \end{minipage}
    \caption{Qualitative comparisons between AFM++~\cite{RegionalAttraction}, F-Clip~\cite{FClip}, LETR~\cite{LETR}, our HAWPv1~\cite{HAWP} and HAWPv2 on the Wireframe dataset~\cite{wireframe-dataset} (the top two rows) and the YorkUrban dataset~\cite{Denis2008} (the last two rows).
    \vspace{-4mm}
    }
    \label{fig:results-display-fsl}
\end{figure*}

\subsubsection{The Results}
We present the quantitative evaluation results using metrics including sAP$^5$, sAP$^{10}$, sAP$^{15}$, AP$^H$, and $F^H$ for detected line segments, as well as mAP$^J$ for detected junctions (Tab.~\ref{tab:summary-metric}). Precision-recall curves for sAP$^5$ and heatmap-based metrics are plotted in Fig.~\ref{fig:pr-curves}.

Compared to L-CNN~\cite{L-CNN} and HAWPv1~\cite{HAWP}, our proposed HAWPv2 achieves significant improvements in challenging metrics such as sAP with varying strictness. HAWPv2 outperforms HAWPv1 by 3.2 and 2.7 points in sAP$^5$ on the Wireframe dataset and the YorkUrban dataset, respectively. Compared to F-Clip~\cite{FClip} and ELSD~\cite{ELSD}, both employing direct regression methods with the Stacked Hourglass Network~\cite{StackedHourglassNetwork} backbone, HAWPv2 outperforms them by at least 3 points in sAP$^5$, the strictest metric. Despite F-Clip and ELSD using larger backbones or advanced design techniques, HAWPv2 still surpasses them. Moreover, HAWPv2 requires significantly fewer training epochs, being 5.67 times less than ELSD and 10 times less than F-Clip.

In terms of heatmap-based evaluation, HAWPv2 demonstrates superior performance on the Wireframe and YorkUrban datasets compared to other approaches. However, AFM++~\cite{RegionalAttraction} remains the best approach with F$^H$ scores of 82.8 and 66.8 on these two datasets.

We also conducted a qualitative comparison between HAWPv2, AFM++~\cite{RegionalAttraction}, HAWPv1~\cite{HAWP}, F-Clip~\cite{FClip}, and LETR~\cite{LETR}. As shown in Fig.~\ref{fig:results-display-fsl}, HAWPv2 achieves overall better results compared to other methods. F-Clip occasionally produces erroneous line segments (\eg, second row), while HAWPv1 and HAWPv2 exhibit greater stability. HAWPv2, with improved visualization accuracy, outperforms HAWPv1 in sAP metrics. While AFM++ does not achieve competitive sAP scores due to its heuristic post-processing, it still produces good visualization results, albeit with localization issues in line segment endpoints.

Regarding inference speed, HAWPv2 achieves 40.8 FPS on a single NVIDIA V100 GPU. The improved inference speed is primarily attributed to our HAT-driven design for proposal generation and verification, which replaces the computationally heavier LOIPooling on the high-dimensional feature map for a fewer number of high-quality line segment proposals. 
\subsection{Ablation Studies}
In this section, we comprehensively evaluate the designs of the line segment proposal generation component (HAT field learning in Sec.~\ref{sec:learning-hat} and endpoint binding in Sec.~\ref{sec:binding}), as well as the line segment proposal verification component (EPD LOIAlign module) in Sec.~\ref{sec:loi-pooling}. All ablation experiments utilize the Wireframe dataset for training and testing, following the settings described in Sec.~\ref{sec:fsl-setting}. The challenging metrics of sAP$^5$, sAP$^{10}$, and sAP$^{15}$ are used for evaluation.

Overall, the EPD LOIAlign module plays the major role in achieving the final performance gain of our HAWPv2 compared to the baselines. Endpoint error loss (Eqn.~\ref{eq:linseg_loss}) in the HAT field learning and the binding threshold choices ($\tau_{\delta}$ in Sec.~\ref{sec:binding}) play a minor role in terms of final accuracy. However, they do play the main role in achieving the accuracy of the proposal, which significantly reduces the number of proposals for verification by 20\%, and thus boosts the inference speed. The expressive power of both EPD LOIAlign and the differentiable HAT field end-point error loss ensures overall effectiveness and efficiency in a disentangled way.  

\subsubsection{The Design of the End-Point Decoupled LOIAlign} \label{sec:design_loialign}
We evaluate the effectiveness of the proposed EPD LOIAlign by comparing it to the baseline, vanilla LOIPooling~\cite{L-CNN}. We analyze three aspects: (1) endpoint-decoupled feature aggregation versus uniform feature aggregation; (2) inclusion or exclusion of features from intermediate points on the HAT field proposed line segments ($\mathbf{Z}_{\psi_\mathbf{x}}(\hat{\ddot{l}}_j)$); and (3) presence or absence of the auxiliary score loss (Eqn.~\ref{eq:ver_aux}).

The results in Tab.~\ref{tab:ablation-dloi-pooling} show that the EPD LOIAlign improves performance by 1.3, 1.1, and 1.0 points for sAP$^{5}$, sAP$^{10}$, and sAP$^{15}$, respectively. Additionally, incorporating features $\mathbf{Z}_{\psi_\mathbf{x}}(\hat{\ddot{l}}_j)$ in line segment verification yields positive effects. The auxiliary classification task aids in learning better features $\mathbf{Z}_{\psi_\mathbf{x}}(\hat{\ddot{l}}_j)$, resulting in performance improvements of 0.3, 0.5, and 0.5 points for sAP$^{5}$, sAP$^{10}$, and sAP$^{15}$, respectively.
In App.~\ref{appx:thin-feature-study}, we compare different feature map dimension $C_{\psi}$ used for representing intermediate points in the EPD LOIAlign, and thus set $C_{\psi}=4$ in our final models.

\begin{table}[]
    \centering
    \caption{The ablation study for the proposed EPD LOIAlign module.}
    \vspace{-3mm}
    \resizebox{0.99\linewidth}{!}{
    \begin{tabular}{c|c|c|c|c|c}
    \toprule
    \makecell{End-Point \\ Decoupled} & \makecell{Features of $\mathbf{Z}_{\psi_\mathbf{x}}(\hat{\ddot{l}}_j)$ \\Initial Lines} & \makecell{ AuxScore (Eqn.~\ref{eq:ver_aux}) \\ BCE Loss} & sAP$^5$ & sAP$^{10}$ & sAP$^{15}$ \\\midrule
    No         &   No   &   No      & 64.0              & 68.0              & 69.8  \\
    Yes        &   No   &   No      & 65.3              & 69.1              & 70.8  \\
    Yes        &   Yes  &   No      & 65.4              & 69.2              & 70.8  \\
    Yes        &   Yes  &   Yes     & \textbf{65.7}     & \textbf{69.7}     & \textbf{71.3}  \\
    \bottomrule
    \end{tabular}
    }
    \vspace{-5mm}
    \label{tab:ablation-dloi-pooling}
\end{table}

\subsubsection{The Designs of Learning Line Segment Proposals}
Line segment proposal quality is crucial for both accuracy and efficiency in wireframe parsing. We conduct detailed experiments on four aspects: (1) EPE loss (Eqn.~\ref{eq:linseg_loss}) and binding threshold $\tau_{\delta}$ (Sec.~\ref{sec:binding}) in (2) training and (3) testing, respectively. Additionally, we verify (4) the effectiveness of residual learning for distance maps and rectified distance maps (Eqn.~\ref{eq:decoding-res}). Throughout these experiments, the EPD LOIAlign component remains unchanged. We compare overall accuracy performance and the average number of line segment proposals per image during inference. {Additionally, we provide references to the average inference latency and breakdown profiling for binding and scoring, allowing further insight into the performance of the system.}

Regarding the EPE loss and $\tau_{\delta}$ settings, we observe that they do not significantly affect the final performance, as shown in Row 6 of Tab.~\ref{tab:ablation-proposals}. This is due to the expressive power of our EPD LOIAlign verification. However, the efficiency drops significantly due to the increased number of line segment proposals (6.19k vs. 2.24k in Row 1). This indicates that HAWPv2 maintains consistently high recall rates of line segments at the proposal generation stage and achieves significant precision improvements with EPE loss.

\begin{table}[]
    \centering
    \caption{The ablation study for the EPE loss and the specification of the line segment binding in training and testing.}
    \vspace{-3mm}
    \resizebox{0.99\linewidth}{!}{
    \begin{tabular}{c|ccc|c|c|c|c}
    \toprule
    &\makecell{EPE Loss \\ (Eqn.~\ref{eq:linseg_loss})}         & \makecell{$\tau_{\delta}$\\(train)} & \makecell{$\tau_{\delta}$\\ (test)} & sAP$^5$ & sAP$^{10}$ & sAP$^{15}$ & \# Proposals\\\midrule
    1 & Yes         &  ${10}$ & ${10}$ & \textbf{65.7}    & \textbf{69.7}       & \textbf{71.3}   & 2.24k    \\
    2 & Yes         &  ${10}$ & $\infty$ & 65.1    & 69.2       & 70.8   & 4.73k    \\\midrule
    3 & Yes         &  $\infty$    & ${10}$ & 65.5 & 69.4 & 71.2   & 2.22k\\
    4 & Yes         &  $\infty$    & $\infty$ &  65.7     & 69.6      &  71.3 & 4.61k\\\midrule
    5 & No          &  $\infty$    & ${10}$ & 65.3   & 69.4  & 71.1 & 2.79k \\
    6 & No          &  $\infty$    & $\infty$    & 65.5   & 69.5  & 71.2 & 6.19k \\
    \midrule
    7 & No          &  ${10}$    & ${10}$ & 65.6 & 69.5 & 71.2 & 2.83k\\
    8 & No          &  ${10}$    & $\infty$ & 65.1 & 69.0 & 70.7 & 6.32k\\
    \bottomrule
    \end{tabular}
    }
    \label{tab:ablation-proposals}
\end{table}

\vspace{0.2em}\noindent \textit{Residual Learning of Distance Maps.} 
{
In Tab.~\ref{tab:ablation-residuals}, not learning distance residuals (Row 1) yields sAP scores of $\{61.1, 65.2, 67.1\}$ for different evaluation strictnesses. However, learning distance residuals as auxiliary supervision signals improves sAP scores by an average of 1.8 points across evaluation thresholds (Row 2), indicating enhanced accuracy in distance map learning. Using the learned distance residuals to generate three rectified distance maps (Eqn.~\ref{eq:decoding-res}) significantly improves performance by an average of 2.17 points (Row 3). For the most strict metric, sAP$^5$, learned distance residuals boost performance by 2.3 points. Modulating the residual scales by adding the modulated distance residuals with $2\times$ scales (-2 and +2) leads to the best precision for our HAWPv2 (Row 4).

Additionally, we compare our strategy for learning unsigned distance residuals with the signed version. We find that signed residual learning only marginally affects performance for different $k$ values, while unsigned residual learning accurately captures the residual distance predictions despite the unknown sign. Unsigned residual learning can be seen as a form of uncertainty estimation rather than regression, while signed residual learning aligns with distance field prediction.

}

\begin{table}[]
    \centering
    \caption{Ablation study evaluating the impact of residual learning on distance maps and scale modulation during the testing phase, including average inference latency and breakdown profiling on a single V100 GPU for binding and scoring.}
    \vspace{-3mm}
    \resizebox{\linewidth}{!}{
        \begin{tabular}{c|c|cccc|cccc}
        \toprule
        \makecell{Learning \\ Distance Residuals} & \makecell{\# Residual\\ Scales} & sAP$^5$ & sAP$^{10}$ & sAP$^{15}$ & \# Proposals & Overall Latency & Binding & Scoring \\\midrule
         No      &  N/A               & 61.1   & 65.2  & 67.1  & 894.84 & 21.821ms & 5.948ms & 1.876ms \\\midrule
        \multirow{3}*{Unsigned}     &  0                 & 62.9   & 67.1  & 68.8  & 931.35 & 22.096ms & 6.046ms & 2.016ms\\
              &  \{-1,0,1\}        & 65.2   & 69.2  & 70.9  & 1704.36& 23.261ms & 7.117ms & 2.335ms\\
              &  \{-2,-1,0,1,2\}   & 65.7   & 69.7  & 71.3  & 2240.09& 24.571ms & 8.065ms & 2.515ms
\\\midrule
        \multirow{3}*{Signed}     &  0                 & 62.6   & 66.3 & 68.2  & 979.78 & 21.915ms & 5.979ms & 2.036ms\\
              &  \{0,1\}        & 62.7 & 66.8 & 68.6 & 995.13 & 22.568ms & 6.625ms & 2.036ms\\
              &  \{0,1,2\}     & 62.7 & 66.9 & 68.6 & 1010.23 &  23.047ms & 6.990ms & 2.043ms\\\bottomrule     
        \end{tabular}
    }
    \vspace{-5mm}
    \label{tab:ablation-residuals}
\end{table}

\section{Experiments of HAWPv3}\label{sec:ssl-exp}
In this section, we test our HAWPv3 under the SSL setting. We also show the out-of-distribution (OOD) capability and potential of our HAWPv3. 

\subsection{Evaluation Protocol and Metrics}

We follow the evaluation protocol presented in SOLD$^2$~\cite{SOLD2}. In detail, we use the repeatability and the localization error as the main metrics across the original input images and the warped ones by the randomly generated homographies. The images in the Wireframe dataset~\cite{wireframe-dataset} and YorkUrban dataset~\cite{Denis2008} are used for evaluation. 

\paragraph*{Repeatibility Scores and Localization Errors} The metrics of repeatability and localization error were extensively used for keypoint detectors and line segment detectors. Given a pair of input images $I$ and $I' = \text{Warp}(I|H)$, where $\text{Warp}(\cdot|H)$ is a homographic image warping function with homography $H\in\mathbb{R}^{3\times 3}$, the repeatability score is calculated by checking if a line segment $\lseg$ in the image is successfully detected again in the warped image up to a distance metric. Denoted by the line segment $\lseg = (\mathbf{x}_1,\mathbf{x}_2)$ and the re-detected one $\lseg' = (\mathbf{x}_1',\mathbf{x}_2')$, the structural distance (\ie, the Euclidean endpoint distance)
\begin{equation}\label{eq:structural-distance}
\begin{split}
    d_{s}(\lseg, \lseg') = \frac{1}{2}\min( & \|\mathbf{x}_1 - \mathbf{x}_1'\|_2 + \|\mathbf{x}_2 - \mathbf{x}_2'\|_2, \\
                  & \|\mathbf{x}_1 - \mathbf{x}_2'\|_2 + \|\mathbf{x}_2 - \mathbf{x}_1'\|_2 ),
\end{split}
\end{equation}
and the orthogonal distance
\begin{equation}\label{eq:orthogonal-distance}
\begin{split}
    d_{orth}(\lseg, \lseg') =\frac{1}{2} (& \|\mathbf{x}_1 - p_{\lseg'}(\mathbf{x}_1)\|_2 + \|\mathbf{x}_2 - p_{\lseg'}(\mathbf{x}_2)\|_2 \\
                         +  & \|\mathbf{x}_1' - p_{\lseg}(\mathbf{x}_1)\|_2 + \|\mathbf{x}_2' - p_{\lseg}(\mathbf{x}_2')\|_2),
\end{split}
\end{equation}
are used to compute the repeatability scores. In the orthogonal distance metric $ d_{o}(\lseg, \lseg')$, the function $p_{\lseg'}(\mathbf{x}_1)$ orthogonally maps the endpoint $\mathbf{x}_1$ into the line segment $\lseg'$. 

Using the distance metrics $d_s$ and $d_{orth}$, we calculate the repeatability scores by counting the co-detected line segments in the image pair $(I,I')$ and its swapped counterpart $(I',I)$ among all detected line segments in the first image of the input pair. A distance threshold $\varepsilon$ of 5 pixels is used. The localization error is then obtained by averaging the distance values over the pairly-detected line segments. We denote the repeatability and localization error for distance threshold $\varepsilon$ as $\text{Rep}_\varepsilon$ and $\text{Loc}_\varepsilon$, respectively.
In contrast to the fully-supervised learning evaluation protocol that resizes predictions to a $128\times 128$ image domain, Eq.~\eqref{eq:structural-distance} and Eq.~\eqref{eq:orthogonal-distance} employ the $\ell_2$ norm (without squaring) on the same domain as the input images for SSL evaluation.

\paragraph*{Random Homography Generation} 
We adopt the homography configurations from SOLD$^2$ to compute repeatability scores and localization errors. The random homography generator takes a patch ratio of 0.85 as input. The perspective displacement, left horizontal displacement, and right horizontal displacement values are obtained using a Gaussian noise generator with a perspective amplitude of 0.2 in both $x$ and $y$ directions, twice the standard deviation. The scaling matrix follows a zero mean Gaussian distribution with a standard deviation of 0.1. Random translations follow a uniform distribution within valid areas, and the rotation matrix follows a uniform distribution with rotation angles ranging from $-\pi/2$ to $\pi/2$.
\begin{figure}
    \centering
    \resizebox{\linewidth}{!}{
        \begin{tabular}{ccc}
             \includegraphics[width=0.31\linewidth]{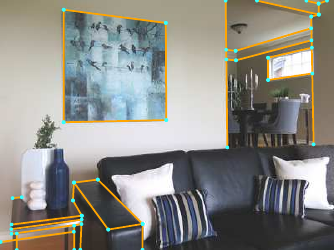} &  
             \includegraphics[width=0.31\linewidth]{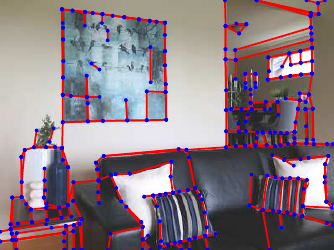} &
             \includegraphics[width=0.31\linewidth]{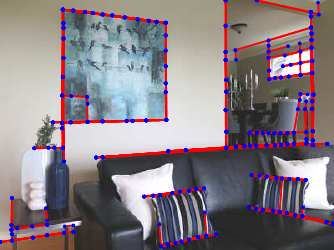}
             \\
             \includegraphics[width=0.31\linewidth]{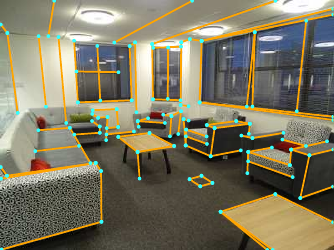} &  
             \includegraphics[width=0.31\linewidth]{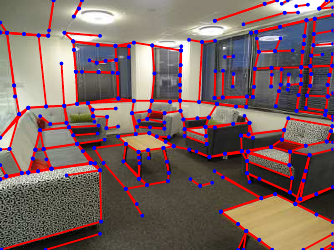} &
             \includegraphics[width=0.31\linewidth]{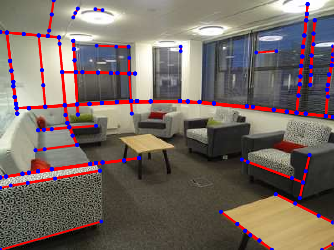}
             \\
             \makecell{HAWPv2@0.95\\\emph{(26/68)}} & \makecell{HAWPv3@0.05\\\emph{(405/404)}} & \makecell{SOLD$^2$~\cite{SOLD2}@0.05\\\emph{(642/443)}}
        \end{tabular}
    }
    \caption{Parsing result comparisons between HAWPv2, HAWPv3 and SOLD$^2$~\cite{SOLD2}. The numbers in the brackets are the number of parsed line segments for the two images, respectively, by each method. }
    \vspace{-5mm}
    \label{fig:fsl-vs-ssl}
\end{figure}

\paragraph*{Datasets} Similar to the experiments of the FSL pipeline, we use the 462 test images of the Wireframe dataset and the 102 images of the YorkUrban dataset as image sources. For each original image, we generate two homographies and independently evaluate the repeatability scores and localization errors on these two datasets. So, there are 924 and 204 pair of images in the Wireframe and YorkUrban datasets for evaluation.

\begin{table*}[!t]
    \centering
    \caption{
        The repeatability evaluation results. Numbers with {\bf bold font} and \underline{underline} indicate the best and second best performance on specific metrics.
        The image resolutions are fixed to $512\times 512$ for both Wireframe and YorkUrban datasets in evaluation.
    }
    \label{tab:exp-repeatibility}
    \vspace{-3mm}
    \resizebox{0.95\linewidth}{!}{
    \begin{tabular}{r|ccccc|ccccc}
    \toprule
    \multirow{3}{*}{Method}     & \multicolumn{5}{c}{Wireframe Dataset} & \multicolumn{5}{c}{YorkUrban Dataset} \\
         \cmidrule{2-11}
         & \multicolumn{2}{c}{$d_s$} & \multicolumn{2}{c}{$d_{\text{orth}}$} &
         \multirow{2}{*}{\#lines / image}
         & \multicolumn{2}{c}{$d_s$} & \multicolumn{2}{c}{$d_{\text{orth}}$} & \multirow{2}{*}{\#lines / image}
         \\
         & Rep-5 $\uparrow$ & Loc-5 $\downarrow$ & Rep-5 $\uparrow$ & Loc-5 $\downarrow$
         & & Rep-5 $\uparrow$ & Loc-5 $\downarrow$ & Rep-5 $\uparrow$ & Loc-5 $\downarrow$
         \\
         \cmidrule{1-11}
    L-CNN~\cite{L-CNN}@0.98     & 0.434 & 2.589 & 0.570 & 1.725 & 76 & 0.318 & 2.662 & 0.449 & 1.784 & 103\\
    DeepHough~\cite{DeepHoughPrior}@0.9     &  0.419 & 2.576 & 0.618 & 1.720 & 135 & 0.315 & 2.695 & 0.535 & 1.751 & 206\\
    TP-LSD~\cite{TP-LSD}TP512 & 0.547 & 2.479 & 0.695 & 1.474 & 77 & 0.447 & 2.491 & 0.610 & 1.491 & 130\\\cmidrule{1-11}
    LSD~\cite{LSD} & 0.383 & 2.198 & 0.719 & 1.028 & 441 & 0.419 & 2.123 & 0.723 & 0.959 & 591\\
    SOLD$^2$~\cite{SOLD2} w/CS & 0.566 & 2.039 & 0.805 & 1.135 & 116 & 0.585 & 1.918 & 0.824 & 1.097 & 196\\
    SOLD$^2$~\cite{SOLD2}  & \underline{0.613} & \underline{2.060} & \textbf{0.921} & \underline{0.809} & 482.6 & 0.629 & 1.951 & {\bf 0.939} & {\bf 0.693} & 1031\\\cmidrule{1-11}
    HAWPv1~\cite{HAWP}@0.97 & 0.451 & 2.625 & 0.537 & 1.738 & 47 & 0.295 & 2.566 & 0.368 & 1.757 & 59\\
    HAWPv2 (Ours)@0.9       & 0.514          & 2.375          & 0.577             & 1.548          & 34  & 0.385 & 2.205 & 0.425 & 1.397 & 30\\
    HAWPv3 (Ours)@0.5       & \textbf{0.751} & \textbf{1.487} & \underline{0.874} & \textbf{0.841} & 145 & {\bf 0.711} & {\bf 1.454} & \underline{0.829} & \underline{0.839} & 225\\
    \bottomrule
    \end{tabular}
    }
\end{table*}

\subsection{Comparisons with the State of the Arts}\label{sec:ssl-evaluation}
We comprehensively compare our proposed HAWPv3 model with the traditional LSD~\cite{LSD}, the fully supervised approaches including L-CNN~\cite{L-CNN}, DeepHough~\cite{DeepHoughPrior}, TP-LSD~\cite{TP-LSD}, HAWPv1~\cite{HAWP} and HAWPv2, as well as the state-of-the-art SSL method, SOLD$^2$~\cite{SOLD2}.  

Tab.~\ref{tab:exp-repeatibility} reports the results of the quantitative comparison. In terms of the structural distance metric $d_s$ defined in Eq.~\eqref{eq:structural-distance}, our HAWPv3 model improves the Rep-5 repeatability scores by 13.8 points at most while obtaining a localization error of 1.487 pixels on the Wireframe dataset compared to the state-of-the-art SOLD$^2$ without the candidate selection scheme. 
For the model SOLD2 with a candidate selection scheme (w/ CS), our HAWPv3 model obtains an improvement by about 20 points. 

In terms of the orthogonal distance metric $d_{orth}$, as it is friendly to the overlapped detection results, the repeatability score of our HAWPv3 is worse than SOLD$^2$, placing it in the second place among all state-of-the-art approaches. For the localization error with $d_{orth}$, our HAWPv3 is still the best. Similar conclusions can be drawn from the YorkUrban dataset. Benefitting from the fast convergence speed of our HAWPv3, we are able to train it within 24 hours on a single GPU (Nvidia RTX 3090) from scratch to obtain very competitive performances. For detailed analysis of our HAWPv3, please refer to Appx.~\ref{appx:detailed-hawpv3}.

\vspace{0.2em}\noindent\textit{Remarks on the FSL and SSL of Wireframe Parsing.}  
For the FSL pipeline, wireframes are typically annotated in a viewpoint-specific manner based on given images, often consisting of long line segments. This poses a challenge for FSL wireframe parsers to achieve high repeatabilities across (warped) views, as viewpoint occlusions can break long line segments into shorter visible ones (see results in Tab.~\ref{tab:exp-repeatibility}). To visually demonstrate this, we compare parsing results in Fig.~\ref{fig:fsl-vs-ssl} between our HAWPv2, HAWPv3, and SOLD$^2$. It is evident that many long line segments detected by HAWPv2 are "split" into several short line segments by HAWPv3 and SOLD$^2$. This difference highlights two key observations: (1) Applying an FSL wireframe parser directly to multiview tasks requiring correspondences across viewpoints can be more challenging compared to SSL wireframe parsers, and (2) SSL wireframe parsers should explore stronger SSL pretext tasks and more effective inductive biases that are learnable and transferable from simulation to reality, in addition to learned edge maps, to approach human perception of line segments.
Between our HAWPv3 and the SOLD$^2$~\cite{SOLD2} at a lower threshold of 0.05, our HAWPv3 can cover more geometry structures in the images  than SOLD$^2$, while using a less number of, but longer, line segments. SOLD$^2$ often computes many overlapped co-linear line segments with different end-points, which contributes to its higher repeatability score, $d_{orth}$.

\begin{figure}[!tb]
	\centering
	\begin{tabular}{ccc}
	\includegraphics[width=0.25\linewidth]{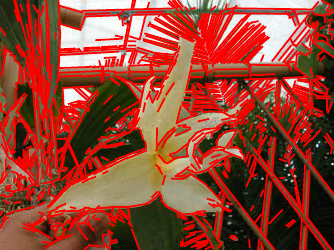}&
    \includegraphics[width=0.25\linewidth]{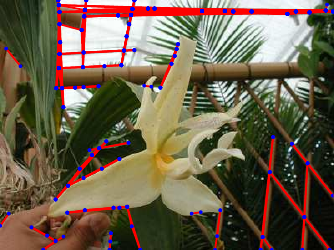}&
    \includegraphics[width=0.25\linewidth]{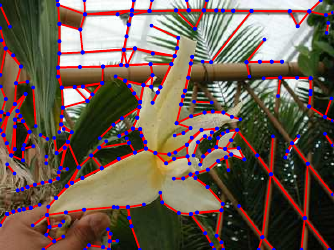}
	\\
    \includegraphics[width=0.25\linewidth]{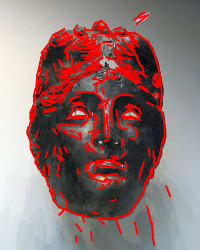}&
	\includegraphics[width=0.25\linewidth]{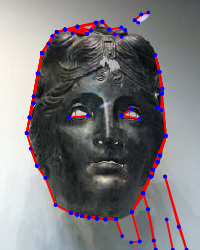}&
	\includegraphics[width=0.25\linewidth]{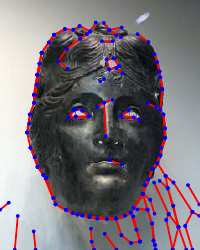}
    \\
    \includegraphics[width=0.25\linewidth]{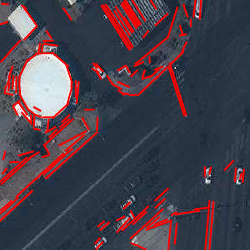}&
    \includegraphics[width=0.25\linewidth]{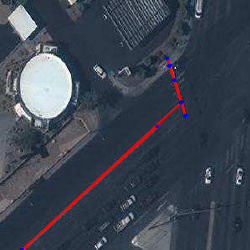}&
    \includegraphics[width=0.25\linewidth]{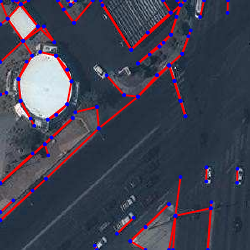}
	\\
    \includegraphics[width=0.25\linewidth]{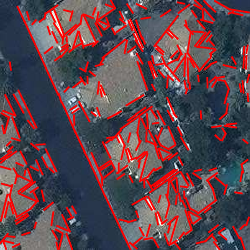}&
    \includegraphics[width=0.25\linewidth]{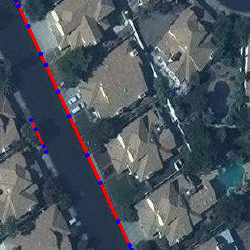}&
    \includegraphics[width=0.25\linewidth]{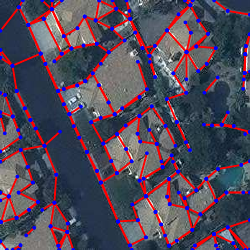}\\
    LSD~\cite{VonGioi2010} & SOLD$^2$~\cite{SOLD2} & HAWPv3 (Ours)
	\end{tabular}
    \caption{Qualitative OOD comparisons between the LSD~\cite{VonGioi2010}, the SOLD$^2$~\cite{SOLD2} and our HAWPv3. The first two rows show results on images in the ImageNet~\cite{ImageNet} dataset, and the last two rows show results on images in the AICrowd~\cite{AICrowdDataset} dataset. 
    \vspace{-5.5mm}
    }
    \label{fig:generalization}

\end{figure}

\subsection{The OOD Potential of our HAWPv3}
To achieve robust geometry understanding of the visual world, it is crucial to possess out-of-distribution (OOD) perception generalizability for low-level structures like line segments and junction points. We qualitatively tested this aspect and observed that our HAWPv3 model exhibits remarkable OOD potential (Fig.~\ref{fig:generalization}). We compared it with the training-free LSD method~\cite{LSD} and the SOLD$^2$ approach~\cite{SOLD2}. By utilizing SSL models trained on the wireframe dataset~\cite{wireframe-dataset} and evaluating them on images from the ImageNet dataset~\cite{ImageNet} and the AICrowd dataset~\cite{AICrowdDataset}, which differ significantly from the indoor images in the wireframe dataset, we observed promising OOD results with our HAWPv3 model. These results offer valuable insights for the development of geometry-guided semi-supervised learning (SSL) approaches for downstream tasks such as image classification and object detection, as well as multi-view 3D vision problems including but not limited to Structure from Motion and SLAM.

\section{Conclusion}\label{sec:conclusion}

This paper comprehensively studies the problem of wireframe parsing from the perspective of line segment representation, the algorithmic design of wireframe parsing, as well as the learning of wireframes in both supervised and self-supervised pipelines. With the proposed novel HAT field representation of line segments that has built-in geometry-awareness, context-awareness and robustness, the presented HAWP models set several new records on challenging benchmarks of Wireframe and YorkUrban while being efficient for training and inference. For the fully-supervised learning with precisely-annotated wireframe labels, the proposed HAWPv2 model is leading the performance on the Wireframe and YorkUrban datasets. For self-supervised learning, the proposed HAWPv3 model shows its advantages of learning the appropriate inductive biases of line segments via the HAT field representation. Besides the strong performance on the structural repeatability, the proposed HAWPv3 model is of great potential for general wireframe parsing in images out of the training distributions.

\ifCLASSOPTIONcaptionsoff
  \newpage
\fi

\section*{Acknowledgements}
{\small 
We acknowledge the efforts of the authors of the Wireframe dataset and the YorkUrban dataset. These datasets make accurate line segment detection and wireframe parsing possible. We thank the anonymous reviewers and associate editors for their constructive comments. We also thank Rémi Pautrat for helpful discussions, and Bin Tan for proofreading. This work was supported by the NSFC Grants under contracts No. 62101390, 62325111 and U22B2011. 
T. Wu was partly supported by ARO Grant W911NF1810295, NSF IIS-1909644, ARO Grant W911NF2210010, NSF IIS-1822477, NSF CMMI-2024688 and NSF IUSE-2013451.  
This work was also supported by the EPSRC grant: Turing AI Fellowship: EP/W002981/1.
The views presented in this paper are those of the authors and should not be interpreted as representing any funding agencies.
}

\bibliographystyle{IEEEtran}
\bibliography{references}

\appendices

\section{Network Architectures}\label{appx:networks}
To facilitate real-time inference speed, we utilize light-weight neural building blocks for different components in the HAWP models. We summarize them as follows. More details are referred to our open-sourced code.  

\begin{itemize}
    \item For the image feature backbone $f_b(\cdot; \Omega_b)$, we use the Stacked Hourglass Networks~\cite{StackedHourglassNetwork} (2 stacks). The dimension of the output feature map $F$ is $C=256$, and the overall stride $s=4$. 
    \item  In learning the HAT field (Sec.~\ref{sec:learning-hat}), we use a vanilla convolution block configuration for $f_d(F; \Omega_d)$ (Eqn.~\ref{eq:distance}),  $f_a(F; \Omega_a)$ (Eqn.~\ref{eq:angle}) and $f_{\Delta d}(F;\Omega_{\Delta d})$ (Eqn.~\ref{eq:dresidual}). 
    \begin{align}
    f_d, f_{\Delta d}: \quad & \sigma\circ\text{Conv}_{1\times1}^{D\to 1}\circ\text{ReLU}\circ\text{Conv}_{3\times3}^{C\to D}, \label{eq:dis-pred}\\
    f_a: \quad & \sigma\circ\text{Conv}_{1\times1}^{D\to 3}\circ\text{ReLU}\circ\text{Conv}_{3\times3}^{C\to D}, \label{eq:angles-pred}    
    \end{align}
    where $\sigma(\cdot)$ is the sigmoid function, $\text{Conv}_{k\times k}^{u\to v}$ denotes the convolution layer that transforms a $u$-channel feature map to a $v$-channel one using a $k\times k$ kernels. In our experiments, we set $D=128$. 
    \item In learning the heatmap-offset field of endpoints (Sec.~\ref{sec:learning-juncs}),  $f_{pt}(F; \Omega_{pt})$ (Eqn.~\ref{eq:endpt}) and $f_{o}(F; \Omega_{o})$ (Eqn.~\ref{eq:offset}) are implemented by $\text{Conv}_{1\times 1}^{C\to 1}$ and $\text{Conv}_{1\times 1}^{C\to 2}$ respectively. 
    \item In computing features for the intermediate points on a line segment proposal in the proposed EPD LOIAlign (Sec.~\ref{sec:loi-pooling}), $f_{\psi}(F;\Omega_{\psi})$ is implemented by $\text{ReLU}\circ\text{Conv}_{3\times3}^{C\to C_{\psi}}$, where $C_{\psi}=4$ in our experiments. 
    \item The MLPs used in the verification head classifier (Eqn.~\ref{eq:ver}) are implemented by two hidden layers with the ReLU nonlinearity function.  
\end{itemize}

\section{The Low-Dimensional Feature Maps $F_{\psi}$ in the EPD LOIAlign Verification}\label{appx:thin-feature-study}
In this ablation study following those in Sec.~\ref{sec:design_loialign}, we show the performance differences in terms of the dimensions $C_{\psi}$'s of computing the ``thin" feature maps $F_{\psi}$. As reported in Tab.~\ref{tab:ablation-dloi-thin-feature}, when the number of feature channels is set to $C_{\psi}=4$, our proposed HAWPv2 model obtains the best performance in terms of accuracy. For the model that uses $C_{\psi}=1$, the performance of accuracy is degenerated by 1.6\% in terms of sAP$^5$. When we increase the channels from $4$ to $8$, there is no  performance gain in both aspects of accuracy and speed. Overall, since the used feature maps in comparisons are all sufficiently ``thin", the inference speed change with respect to the number of feature channels is less apparent.

\begin{table}[!h]
    \centering
    \caption{The performance change by the different number of feature channels for the thin feature maps.}
    \label{tab:ablation-dloi-thin-feature}
    \begin{tabular}{c|ccc|c}
    \toprule
    Feat. Dim. $C_{\psi}$ & sAP$^{5}$ & sAP$^{10}$ & sAP$^{15}$ & FPS\\\midrule
    1          & 64.1 & 68.1 & 69.8 & 42.5 \\
    2          & 64.6 & 68.6 & 70.2 & 41.4 \\
    4          & \textbf{65.7} & \textbf{69.7} & \textbf{71.3} & 40.8 \\
    8          & 65.6 & 69.3 & 71.0 & 40.0 \\
    \bottomrule
    \end{tabular}
\end{table}

\section{More Detailed Analyses of HAWPv3}\label{appx:detailed-hawpv3}

\begin{table} [!ht]
    \centering
    \caption{The quantative evaluation results of the synthetically trained models, SOLD$^2$ and our HAWPv3.
    All the evaluation results are obtained with the detection threshold of 0.5 and inlier threshold of 0.75. 
    }
    \label{tab:exp-repeatibility-spring}
    \resizebox{0.8\linewidth}{!}{
    \begin{tabular}{cc|ccc}
    \toprule
                                &                       & SOLD$^2$ & SOLD$^2$ (w/ CS) & HAWPv3 \\\midrule
    \multirow{2}{*}{{$d_s$}}      & Rep-5 $\uparrow$      & 0.306    & 0.282            &   {\bf 0.356} \\
                                & Loc-5 $\downarrow$    & 2.697    & {\bf 2.636}            &   2.912 \\\cmidrule{3-5}
    \multirow{2}{*}{$d_{orth}$} & Rep-5 $\uparrow$      & 0.573    & 0.384            &   {\bf 0.629} \\                            
                                & Loc-5 $\downarrow$    & {\bf 1.233}    & 1.367            &   1.890 \\\midrule
    \multicolumn{2}{c|}{\#lines/image}                  & 310            & 95         &   170   \\
    \bottomrule                            
    \end{tabular}
    }
    
\end{table}

\paragraph*{Evaluation of Synthetically-Trained Models} As the self-supervised wireframe parsers spring from the synthetically-trained models, we evaluate our HAWPv3 and SOLD$^2$~\cite{SOLD2} before adopting the Homography Adaptation learning scheme on the real-world images. We use the metrics of repeatability scores and localization errors as in Sec.~\ref{sec:ssl-evaluation} on the Wireframe dataset. As reported in Tab.~\ref{tab:exp-repeatibility-spring}, our proposed HAWPv3 model that is trained only using the synthetic data achieves better repeatibility scores for both $d_s$ and $d_{orth}$ distance metrics than SOLD$^2$ and SOLD$^2$ (w/CS). For the localization error, SOLD$^2$ obtains the best on $d_{orth}$ metric and the candidate selection (w/ CS) scheme improves the localization error on the $d_{s}$ metric. Although the localization error by our HAWPv3 is larger than SOLD$^2$, our final model is significantly better than SOLD$^2$ (see Tab.~\ref{tab:exp-repeatibility}).

\paragraph*{Iterative Learning and Refinement (Cascade SSL)} Thanks to the fast pseudo wireframe label generation and the efficient training schedule of our proposed HAWPv3, we have the computational budget for iteratively training the HAWPv3 model with more and more accurate pseudo wireframe labels (\ie, cascade SSL). In this ablation study, we run several experiments to study the possible settings of the iterative learning and refinement: (1) Random model initialization v.s. Warmp-up model initialization. (2) The training schedules in terms of the initial learning rate, the number of epochs and the learning rate decay milestone(s). In the comparisons, we denote the HAWPv3 models by \texttt{HAWPv3}$^{(k)}$ as the model trained with the pseudo labels generated by the \texttt{HAWPv3}$^{(k-1)}$ ($k\geq 1$).  \texttt{HAWPv3}$^{(0)}$ is the  synthetically-pretrained model and used in generating the initial pseduo wireframe labels for real images. The results are summarized in Tab.~\ref{tab:ask-answer}. We have the observations as follows.

\begin{table}[]
    \centering
    \caption{The ablation study on the settings of iteratively training HAWPv3 models. It is done based on the repeatibility and localization performance on the Wireframe dataset. For the training schedule, we denote its initial learning rate, the total training epochs and the decay milestone by \texttt{lr/epochs/milestone}. }
    \label{tab:ask-answer}
    \resizebox{0.97\linewidth}{!}{
    \begin{tabular}{l|ll|ccccc}
    \toprule
    \multirow{2}{*}{\makecell{Model\\ Name}} & \multirow{2}{*}{\makecell{Init.\\ Type}} & \multirow{2}{*}{Sched.} & \multicolumn{2}{c}{$d_s$} & \multicolumn{2}{c}{$d_{orth}$} & \multirow{2}{*}{\makecell{\#lines\\/image}}\\
                        & & & Rep-5$\uparrow$ & Loc-5$\downarrow$ & Rep-5$\uparrow$ & Loc-5$\downarrow$ \\\midrule
                        
    \texttt{HAWPv3}$^{(1)}$\texttt{-A}*   & Random & 4e-4/30/25 & 0.691 & 1.672 & 0.806 & 0.944 & 104  \\
    \texttt{HAWPv3}$^{(1)}$\texttt{-B}    & \texttt{HAWPv3}$^{(0)}$  & 4e-4/30/25 & 0.610 & 1.790 & 0.762 & 0.946 & 101  \\
    \midrule
    \texttt{HAWPv3}$^{(2)}$\texttt{-A}  & Random & 4e-4/30/25 & 0.700 & 1.504  & 0.833 & 0.844 & 147  \\
    \texttt{HAWPv3}$^{(2)}$\texttt{-B}   & \texttt{HAWPv3}$^{(1)}$\texttt{-A} & 4e-4/30/25 & 0.707 & 1.569  & 0.835 & 0.888 & 141\\
    \texttt{HAWPv3}$^{(2)}$\texttt{-C}*   & \texttt{HAWPv3}$^{(1)}$\texttt{-A} & 4e-5/30/25  & 0.751 & 1.487 & 0.874 & 0.841 & 145\\\midrule
    \texttt{HAWPv3}$^{(3)}$\texttt{-A}   & \texttt{HAWPv3}$^{(2)}$\texttt{-C} & 4e-5/30/-  & 0.741 & 1.509 & 0.883 & 0.847 & 166\\
     \texttt{HAWPv3}$^{(3)}$\texttt{-B}   & \texttt{HAWPv3}$^{(2)}$\texttt{-C} & 4e-6/30/25 & 0.739 & 1.503 &  0.879 & 0.844 & 161\\
    \bottomrule
    \end{tabular}
    }
\end{table}

First, at the first stage of the cascade SSL, we do not need to use the synthetically pre-trained model weights to warm up the model to be trained on the real images with the pseudo-wireframe labels. With the same training schedule, \texttt{HAWPv3}$^{(1)}$\texttt{-A} outperforms \texttt{HAWPv3}$^{(1)}$\texttt{-B} by large margins, which can be intuitively understood based on the semantic gap between synthetic images and real images.  Compared to SOLD$^2$ (see Tab.~\ref{tab:exp-repeatibility}),  \texttt{HAWPv3}$^{(1)}$\texttt{-A} obtains an absolute improvement by 9.6 points on the repeatability score.

Second, after the first stage of the cascade SSL, the real image-trained HAWPv3 model can be leveraged in warming up the model weights in the next stage, and better performance can be achieved with a more conservative initial learning rate. With the same training schedule, \texttt{HAWPv3}$^{(2)}$\texttt{-A}  and \texttt{HAWPv3}$^{(2)}$\texttt{-B} have very similar performance. With a reduced initial learning rate, \texttt{HAWPv3}$^{(2)}$\texttt{-C} improves the repeatability scores by 3 points and 4.4 points for the two distance metrics respectively,  while reducing the localization errors by about 14\% for both $d_s$ and $d_{orth}$. Compared with \texttt{HAWPv3}$^{(1)}$ models, the average number of SSL ``annotated" line segments is significantly increased (\eg., 98 vs. 134). Note that the overall training cost of \texttt{HAWPv3}$^{(2)}$\texttt{-C} is still significantly less than that used by the SOLD$^2$~\cite{SOLD2}.

Last but not least, the potential of the cascade SSL is quickly saturated, which is reasonably expected due to the essence of the simulation-to-reality SSL pipeline. On top of the model weights of \texttt{HAWPv3}$^{(2)}$\texttt{-C}, we do not observe further improvement with different training schedules in \texttt{HAWPv3}$^{(3)}$\texttt{-A} and \texttt{HAWPv3}$^{(3)}$\texttt{-B}.  For more training rounds with lower learning rates (\eg, 4e-7), we also did not observe any improvement.  We did not explore the mixed training in which both the synthetic training data and the SSL ``annotated real data are used.

\begin{table}[!t]
    \centering
    \caption{The ablation study on the model architecture designs for self-supervised learning. HAWPv3 inherits the main design of HAWPv2 while HAWPv3$^\dag$ comes from HAWPv1~\cite{HAWP}. }
    \label{tab:ask-answer-2}
    \resizebox{0.97\linewidth}{!}{
    \begin{tabular}{l|c|ccccc}
    \toprule
    \multirow{2}{*}{\makecell{Model\\ Name}} & \multirow{2}{*}{\makecell{Learning Stage}} & \multicolumn{2}{c}{$d_s$} & \multicolumn{2}{c}{$d_{orth}$} & \multirow{2}{*}{\makecell{\#lines\\/image}}\\
                        & & Rep-5$\uparrow$ & Loc-5$\downarrow$ & Rep-5$\uparrow$ & Loc-5$\downarrow$ \\\midrule
                        
    \texttt{HAWPv3}          & \multirow{2}{*}{Synthetic} &  \bf 0.388 & \bf 2.863 & \bf 0.608 & \bf 1.887 & 76  \\
    \texttt{HAWPv3$^\dag$}   &                            &  0.278 & 3.429 & 0.536 & 2.439 & 89  \\\midrule
    \texttt{HAWPv3}          & \multirow{2}{*}{Real (Round 1)} & \bf 0.691 & \bf 1.672 & \bf 0.806 & \bf 0.944 & 104  \\
    \texttt{HAWPv3$^\dag$}   &                            &  0.541 & 3.054 & 0.772 & 1.895 & 178 \\\midrule
    \texttt{HAWPv3}          & Real (Round 2) & \bf 0.751 & \bf 1.487 & \bf 0.874 & \bf 0.841 & 145  \\
    
    \bottomrule
    \end{tabular}
    }
\end{table}

\section{SSL Extensibility for HAWPv1 and HAWPv2} 
Because the architecture of HAWPv3 is primarily based on HAWPv2, which has demonstrated outstanding performance in fitting human annotation wireframe labels, a natural question arises: "Is it necessary to upgrade HAWPv1 to HAWPv2 first in order to obtain HAWPv3 in self-supervised learning?" To quantitatively address this question, we adapt the HAWPv1 model by adding a high-resolution edge map learning branch and train a series of models to evaluate repeatability under the SSL evaluation protocol. We refer to the model adapted from HAWPv1 as HAWPv3$^\dag$.
As shown in Table~\ref{tab:ask-answer-2}, HAWPv3$^\dag$ achieves inferior performance at all learning stages. In the synthetic stage, the repeatability scores on the wireframe dataset decrease by 11 points and 7.2 points in terms of $d_s$ and $d_{\rm orth}$, respectively. During the first round of learning with real-world images, further performance degradation of HAWPv3$^\dag$ is observed compared to HAWPv3, particularly in terms of localization errors. These results indicate that achieving similar or better performance than SOLD$^2$ with the HAWPv1 architecture would be challenging, further justifying our design rationale for developing HAWPv2 to enable self-supervised learning.

\end{document}